\documentclass[journal]{new-aiaa} 
\usepackage[utf8]{inputenc}

\usepackage{graphicx}
\usepackage{amsmath}
\usepackage[version=4]{mhchem}
\usepackage{siunitx}
\usepackage{longtable,tabularx}
\usepackage{subcaption}
\usepackage{makecell}
\usepackage{multirow}
\usepackage{subcaption}
\usepackage{multirow}
\usepackage{booktabs}
\usepackage{url}
\def\thickhline{\noalign{\hrule height.8pt}}
\usepackage{listings}
\usepackage{xcolor}

\newif\ifshowedits
\showeditsfalse 

\ifshowedits
    \newcommand{\edit}[1]{\textcolor{red}{#1}}
\else
    \newcommand{\edit}[1]{#1}
\fi

\title{\edit{Language-Conditioned Safe Trajectory Generation for Spacecraft Rendezvous}}

\author{Yuji Takubo \footnote{Ph.D. Candidate, Department of Aeronautics and Astronautics, 496 Lomita Mall, AIAA Student Member. \edit{Corresponding Author (ytakubo@stanford.edu)}.}, 
Arpit Dwivedi \footnote{Master Student, Department of Aeronautics and Astronautics, 496 Lomita Mall.},
Sukeerth Ramkumar \footnote{Master Student, Department of Mechanical Engineering, 440 Escondido Mall Building 530.}, 
Luis A. Pabon \footnote{Ph.D. Candidate, Department of Aeronautics and Astronautics, 496 Lomita Mall.}, 
Daniele Gammelli \footnote{Postdoctoral Fellow, Department of Aeronautics and Astronautics, 496 Lomita Mall.}, \\ 
Marco Pavone \footnote{Associate Professor, Department of Aeronautics and Astronautics, 496 Lomita Mall. AIAA Associate Fellow.},
Simone D'Amico \footnote{Associate Professor, Department of Aeronautics and Astronautics, 496 Lomita Mall. AIAA Fellow.} 
}
\affil{Stanford University, Stanford, CA, 94305}

\begin{document}

\maketitle

\footnotetext{The preliminary version of this work was presented at the 2026 AIAA SciTech Forum as manuscript AIAA-2026-0350. \cite{takubo2026semantic}.}
\footnotetext{\edit{The codebase supporting the experiments in this paper is publicly available at: \url{https://github.com/UzTak/SAGES}.}}

\begin{abstract}
Reliable real-time trajectory generation is essential for future autonomous spacecraft. 
While recent progress in nonconvex guidance and control is paving the way for onboard autonomous trajectory optimization, these methods still rely on extensive expert input (e.g., waypoints, constraints, mission timelines, etc.), which limits operational scalability in \edit{complex} missions such as rendezvous and proximity operations.
This paper introduces SAGES (Semantic Autonomous Guidance Engine for Space), a trajectory-generation framework that translates natural-language commands into spacecraft trajectories that reflect high-level intent while respecting nonconvex constraints. 
Experiments in two settings (fault-tolerant proximity operations with continuous-time constraint enforcement and a free-flying robotic platform) demonstrate that SAGES reliably produces trajectories aligned with human commands, achieving over 90\% semantic-behavioral consistency across diverse behavior modes.
Ultimately, this work marks an initial step toward language-conditioned, constraint-aware spacecraft trajectory generation, enabling operators to interactively guide both safety and behavior through intuitive natural-language commands with reduced expert burden. 

\noindent Project Website: \url{https://semantic-guidance4space.github.io/}
\end{abstract}

\section{Introduction}
\lettrine{A}{utonomous} decision-making for spacecraft is essential for the emerging era of the future space ecosystem. 
In particular, automated rendezvous, proximity operations, and docking technology enable frequent space logistics operations, including on-orbit servicing, crewed docking, and in-space manufacturing, which could realize a sustainable economy in orbit and beyond. 
A central requirement for these rendezvous missions is real-time trajectory generation that can jointly optimize mission objectives while satisfying complex operational constraints.

Motivated by these ambitious goals, recent advances in nonlinear and nonconvex trajectory optimization have significantly enhanced the autonomy and reliability of spacecraft guidance and control systems.
There exist numerous algorithms that solve optimal control numerically, including direct methods \cite{patterson2014gpops}, indirect methods \cite{lawden1963optimal, koenig2020fast, foss2025efficient}, and differential dynamic programming \cite{gershwin1970discrete}.
Among these approaches, Sequential Convex Programming (SCP) \cite{malyuta_scp_2022} has emerged as one of the most promising nonconvex guidance algorithms for practical deployment.
By iteratively solving a convexified problem until convergence, SCP provides a computationally efficient and reliable strategy for spacecraft trajectory optimization.
Its effectiveness has been demonstrated in a range of aerospace applications, from powered descent and landing \cite{blackmore2016autonomous, strohl2022implementation} to spacecraft rendezvous \cite{berning2024chance, rizza2025goal}.
This practical success is supported by theoretical works establishing convergence guarantees to local optima \cite{mao2016successive, bonalli_2019_gusto, oguri2023successive}, as well as research broadening the class of nonconvex constraints that SCP can handle. 
Notable extensions include mixed-integer formulations \cite{malyuta2023fast, szmuk2020successive, mao2022successive} or continuous-time constraint satisfaction \cite{elango2025continuous}. 
Additional developments address uncertainty through chance-constrained methods \cite{blackmore2011chance} and moment-based techniques such as covariance steering \cite{ridderhof2019nonlinear, kumagai2024sequential, takubo2024multiplicative}.

Despite these advances, the use of nonconvex trajectory optimization in complex rendezvous operations remains limited by two key challenges.
First, human experts must still manually specify problem parameters, such as waypoints, timelines, operational constraints, and objectives.
The ability to \edit{quickly} formulate such problems is essential for autonomous decision-making, responsive operations, and agile mission and navigation analysis. 
Yet achieving this level of flexibility requires deep expertise in optimal control, making large-scale deployment of nonconvex guidance algorithms difficult.
Second, classical objective functions such as fuel or time minimization often lack the expressive power needed to capture the full range of desired behaviors.
As a result, operators rely on the detailed waypoint construction to shape the intended rendezvous trajectory.
However, for space rendezvous, the geometry and kinematics of relative motion \cite{damico_phd_2010, alfriend2009spacecraft} introduce subtleties that require significant astrodynamics knowledge, making it challenging to design waypoints that reliably yield the intended behavior.
Moreover, with the advent of high-capacity propulsion systems for multi-spacecraft servicing and orbit-transfer vehicles, control effort during high-precision proximity operations has become comparatively inexpensive in a wide range of orbital regimes \cite{Guffanti2023Visors, rizza2026space, takubo2025safe}.
This shift unlocks a design space where slight increases in fuel usage can be traded for diverse behaviors and more complex maneuvers, echoing the concept of goal-oriented autonomy \cite{ceballos2011goal, rizza2025goal}.
Nevertheless, current methods based on optimal control formulations that focus solely on cost minimization are inherently limited in their ability to leverage this opportunity.

Given these challenges, a language-conditioned interface offers a promising new direction for trajectory generation in space.
Such an interface would allow non-expert spacecraft operators and systems engineers to specify high-level mission objectives in natural language, facilitating rapid synthesis of dynamically feasible trajectories that align with task semantics.
In this paper, such trajectories are referred to as \textit{semantically correct}.

\smallskip
Concurrently, the emergence of internet-scale, broadly capable Foundation Models (FMs) offers an opportunity to fundamentally rethink how autonomous systems are designed, deployed, and operated. 
Trained on vast and diverse datasets, these models capture broad priors about the world and have achieved breakthroughs in vision, language, and mfultimodal reasoning.
Building on this momentum, a substantial body of work has explored how to adapt pretrained FMs for robot control \cite{lynch2020language, luketina2019survey, kim2024openvla, kawaharazuka2025vision}. 
Within the space domain, Large Language Model (LLM)-driven spacecraft autonomy is beginning to emerge, as demonstrated by systems that directly map textual commands to control inputs \cite{carrasco2025large, carrasco2025visual, zucchelli2025fine} and by approaches that perform semantic reasoning over multimodal inputs \cite{foutter2025space-llava}.
However, these approaches typically assume closed-loop operation and relegate most of the dynamics and constraint handling to lower-level controllers. 
\edit{While effective for many robotic applications with relatively large instantaneous feasible sets, this paradigm is limited in aerospace settings. 
In such cases, safety must be enforced at the trajectory level, and the admissible state–control space is tightly constrained, necessitating planning approaches that explicitly account for dynamics and constraints.}
Moreover, although spacecraft dynamics are known with far greater precision than those of terrestrial robots (e.g., locomotion involving contact forces), current FM-based methods do not make intrinsic use of this analytical structure during policy generation. 
Instead, they rely on data-driven approximations that lack formal verification, limiting their suitability for safety-critical space missions.

As a first step toward addressing these limitations, this paper introduces \edit{SAGES (Semantic Autonomous Guidance Engine for Space)}, a \edit{language}-conditioned spacecraft trajectory generation framework that translates high-level language commands into feasible trajectories under nonconvex constraints.
The proposed approach bridges semantic intent and continuous dynamics through a two-stage architecture:
(i) a multimodal encoder-decoder model embeds language descriptions and constraint specification into a shared latent space and autoregressively generates a trajectory that reflects the semantic intent, and
(ii) a SCP layer then refines this trajectory to enforce principled constraint satisfaction with respect to dynamics and operational constraints.
This framework builds upon the Autonomous Rendezvous Transformer (ART) \cite{art_ieeeaero24}, a transformer-based trajectory generation model originally designed to provide an initial guess (warm-start) to an SCP. 
Unlike previous ART formulations \cite{art_ieeeaero24, art_ral24, takubo2025towards,  celestini2025generalizable, takubo2026agile}, SAGES treats SCP not as a local optimizer but as a projector to the feasible domain: any converged SCP solution strictly satisfies the encoded constraints while preserving semantic and temporal coherence.
By integrating high-level language conditioning with a dynamics-aware decision model, the proposed framework enables coherent, constraint-satisfying trajectory generation that unifies semantic intent and physics-based optimization. 

\smallskip
The contributions of this paper are threefold. 
\edit{First, this paper proposes a new paradigm for semantic spacecraft guidance that explicitly integrates hard safety constraints with soft, high-level behavioral specifications. 
This is achieved by combining a pre-trained language encoder with a transformer-based trajectory generation model (ART), where the input command is encoded into a semantic embedding that conditions the downstream trajectory generation.
Secondly, the proposed SAGES framework integrates a multimodal transformer-based semantic trajectory generator with an SCP-based feasible-set projector, enabling the computation of safe trajectories that satisfy quantitative task requirements while remaining aligned with high-level semantic intent. Finally, the effectiveness of SAGES is demonstrated in two domains: (i) a fault-tolerant spacecraft proximity operation scenario with continuous-time safety constraints, and (ii) hardware experiments on a free-flyer robotic testbed with an embedded GPU system. Across both settings, SAGES successfully demonstrates semantically guided safe autonomy, improving SCP’s algorithmic performance while generating trajectories that execute the intended behavior.}

\smallskip
As an extended and revised version of the preceding work \cite{takubo2026semantic}, this paper additionally reports statistical analysis of the performance of the proposed framework using an embedded GPU platform, as well as formal experiments and analysis of the run-time monitoring framework.  
Leveraging the latent-space projection of the input command, the monitoring architecture is designed to operate in parallel with the SAGES trajectory generation process.  
This design enables the detection and rejection of text commands that are dissimilar to the training data, thereby adding an additional layer of safety to the \edit{proposed semantic} guidance strategy.

The remainder of the paper is organized as follows. 
Section~\ref{sec:background} reviews ART, a transformer-based trajectory generation framework that is foundational for SAGES. 
Section~\ref{sec:methodology} introduces SAGES, a novel method that incorporates constraint-aware semantic trajectory generation. 
Section~\ref{sec:case_study} presents two experimental setups: a free-flying robotic testbed and a spacecraft proximity operation. 
Section~\ref{sec:results} reports the results of the proposed framework, and Section~\ref{sec:conclusion} concludes the paper.

\section{Background: Autonomous Rendezvous Transformer} \label{sec:background}

The Autonomous Rendezvous Transformer (ART) \cite{art_ieeeaero24} is an approach that combines optimization-based and learning-based methods for spacecraft trajectory optimization.
The method entails leveraging high-capacity (namely, transformer-based) neural network models for the solution of discrete-time optimal control problems of the form:
\begin{subequations} \label{eq:ocp}
\begin{alignat}{2}
    \min_{\{\boldsymbol{x}_k\}_{k=1}^{N}, \{\boldsymbol{u}_k\}_{k=1}^{N}} \quad & 
    \mathcal{J} = \sum_{k=1}^{N} j_{k}(\boldsymbol{x}_k,\boldsymbol{u}_k,t_k)  \\
    \text{subject to} \quad 
    & \boldsymbol{x}_{k+1} = \boldsymbol{F}_k (\boldsymbol{x}_{k}, \boldsymbol{u}_k), 
    && \forall k = 1,...,N-1, \label{eq:ocp_con_dyn} \\
    & (\boldsymbol{x}_k, \boldsymbol{u}_k) \in \mathcal{S}_k, 
    && \forall k = 1,...,N, \label{eq:ocp_con_path}
\end{alignat}
\end{subequations}
where $\boldsymbol{x}_k = \boldsymbol{x}(t_k) \in \mathbb{R}^{n_x}$ and $\boldsymbol{u}_k = \boldsymbol{u}(t_k) \in \mathbb{R}^{n_u}$ denote the system state and control input at discrete timestep $k$.
The objective function $\mathcal{J}$ aggregates the stage costs $j_k$ over a fixed horizon $t\in [0,t_f]$ discretized by $N$ steps with the timestep $\Delta t = t_f/(N-1)$. 
Previous works set the total fuel cost as the objective \cite{art_ieeeaero24, takubo2025towards}, although other cost functionals can also be considered.
In addition, $\boldsymbol{F}_k: \mathbb{R}^{n_x} \times \mathbb{R}^{n_u} \rightarrow \mathbb{R}^{n_x}$ defines the discrete-time system dynamics. 

The admissible state–control pairs at each step are constrained within the feasible set $\mathcal{S}_k \subseteq \mathbb{R}^{n_x} \times \mathbb{R}^{n_u}$, which encodes all operational and physical constraints. 
More generally, the feasible region may consist of multiple disjoint constraint sets, expressed as a finite union $
\mathcal{S}_k = \mathcal{S}_{k,1} \cup \mathcal{S}_{k,2} \cup \dots \cup \mathcal{S}_{k,n_c},$
where each $\mathcal{S}_{k,i}$ represents a distinct admissible subset.

The central idea of ART is to train a transformer \edit{\cite{vaswani2017attention}} that predicts future states and control inputs conditioned on past states, actions, and mission context. 
These predictions serve as a high-quality initial guess for the SCP solver in the solution of the problem in Eq.~\eqref{eq:ocp}, thereby accelerating convergence and enabling the solver to obtain more optimal solutions \edit{\cite{art_ieeeaero24, celestini2025generalizable, takubo2025towards, takubo2026agile}}.
A key enabler of ART for robust trajectory generation is the state representation (i.e., tokenization) of the continuous trajectory information to a sequence of states, controls, and corresponding performance metrics. 
The general tokenized trajectory representation is defined as:
\begin{align}
    \tau_{1:N} = \{ \mathcal{P}_1, \boldsymbol{x}_1, \boldsymbol{u}_1, ..., \mathcal{P}_N, \boldsymbol{x}_N, \boldsymbol{u}_N \},
\end{align}
where $\mathcal{P}_k$ is a performance metric vector. 
In particular, two performance metrics are considered in this work as $\mathcal{P}_k = \{r_k, c_k\}$, where $r_k \in \mathbb{R}$ is the \textit{reward-to-go} and $c_k \in \mathbb{N}$ is the \textit{constraint-to-go}, defined as:
\begin{subequations}
    \begin{align}
    r_k & = -\sum_{j=k}^{N} \mathcal{J}(\boldsymbol{x}_j,\boldsymbol{u}_j,t_j) \\
    c_k & = \sum_{j=k}^{N} \mathbf{1}_{\mathcal{S}_j}(\boldsymbol{x}_j, \boldsymbol{u}_j), \quad \mathbf{1}_{\mathcal{S}_k}(\boldsymbol{x}_k, \boldsymbol{u}_k) := \begin{cases}
        1 \quad & \text{if} \ (\boldsymbol{x}_k,\boldsymbol{u}_k) \in \mathcal{S}_k \\
        0 & \text{otherwise}.
    \end{cases} \label{eq:ctg}
\end{align}
\end{subequations}
Together, these metrics quantify the cumulative optimality and feasibility of the remaining trajectory beginning at step $k$.

\section{SAGES: Semantic Autonomous Guidance Engine for Space} \label{sec:methodology}

\begin{figure}
    \centering
    \includegraphics[width=\linewidth]{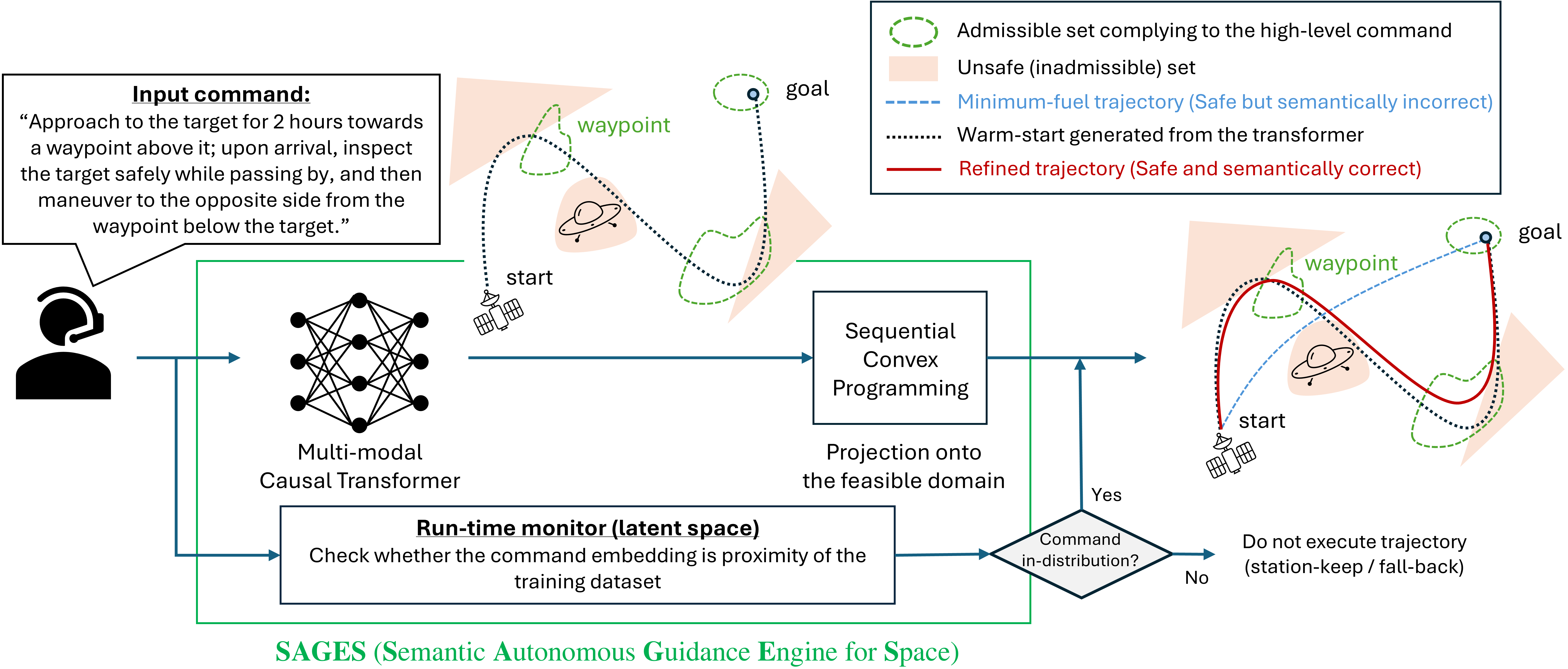}
    \caption{\edit{SAGES (Semantic Autonomous Guidance Engine for Space) generates a language-conditioned safe trajectory via transformer-based warm-start trajectory generation and sequential convex programming with run-time monitoring.}}
    \label{fig:summary}
\end{figure}

Building on ART, this paper introduces SAGES, a framework for language-conditioned trajectory generation. 
\edit{Figure~\ref{fig:summary} illustrates the overall flowchart of SAGES. 
It takes a natural-language command as input and produces a semantically correct, constraint-satisfying trajectory. 
A multimodal encoder–decoder model (a transformer conditioned on text, constraint specifications, and the current system state) autoregressively predicts the optimal next action at each step, generating an initial trajectory that aligns with the requested behavior. 
This trajectory is then refined using SCP, which projects it onto the feasible set. In parallel, the run-time monitor validates whether the input command is in-distribution using latent space information.}

The key distinction between SAGES and ART lies in the role of the transformer. 
Whereas ART predicts a fuel-minimizing trajectory, SAGES uses a transformer to generate trajectories that capture high-level semantic objectives.
SAGES then employs SCP to compute a nearby feasible trajectory, unlike ART, in which SCP is used directly to perform fuel minimization.
This architectural choice ensures that the resulting spacecraft trajectory remains within the feasible domain at all times and prioritizes safety over strict semantic adherence. 
Such a design is particularly well-suited to safety-critical systems such as spacecraft, where safety constraints must be strictly enforced throughout the entire trajectory and operational objectives such as waypoint passage are treated as soft constraints.

\subsection{\edit{Problem Formulation}}

The ultimate goal of SAGES is to generate a trajectory that solves the following problem:
\begin{subequations}
    \begin{align} \label{eq:feas_prob}
    \text{find}  \quad & \{\boldsymbol{x}_k\}_{k=1}^N, \{\boldsymbol{u}_k\}_{k=1}^N \\
    \text{subject to} \quad &  (\{\boldsymbol{x}_k \}_{k=1}^N, \{\boldsymbol{u}_k \}_{k=1}^N) \in  \text{(Set of trajectories consistent with a high-level command)}. \\
    & \eqref{eq:ocp_con_dyn}, \eqref{eq:ocp_con_path}
\end{align}
\end{subequations}
In other words, the generated trajectories must satisfy the hard constraints defined in the optimal control formulation (cf. Eq.~\eqref{eq:ocp}) while also adhering to a soft, semantic constraint that enforces consistency with the behavior specified by the input command.  
This formulation emphasizes feasibility under semantic guidance rather than explicit cost minimization.

To enable \edit{language}-conditioned safe trajectory generation, SAGES adopts a trajectory tokenization scheme that extends ART’s original representation by incorporating text information into a shared latent space \cite{putterman2021pretraining}. 
The resulting token sequence is defined as:
\begin{align} \label{eq:traj_token_sages}
    \tau_{1:N} = \{ \boldsymbol{e}, c_1, \boldsymbol{x}_1, \boldsymbol{u}_1, c_2, \boldsymbol{x}_2, \boldsymbol{u}_2, ..., c_N, \boldsymbol{x}_N, \boldsymbol{u}_N \},
\end{align}
where $\boldsymbol{e}$ denotes the text embedding vector, given by:
\begin{align} \label{eq:txt_encoder}
    \boldsymbol{e} = [\boldsymbol{e}_1, ..., \boldsymbol{e}_{N_e}] = f_{\text{enc}}(z; \theta_{\text{enc}}).
\end{align}
Here, $f_{\text{enc}}$ denotes a text encoder with frozen weights $\theta_{\text{enc}}$, and $z$ is the natural-language input specifying the high-level (trajectory-level) command.
Each encoder output token $\boldsymbol{e}_k \in \mathbb{R}^h$ represents the embedding of a (sub-)word and shares the same dimensionality as the hidden space used for ART’s trajectory tokens, with $h$ being the dimensionality of the latent space.
The full text embedding $\boldsymbol{e}$ has a fixed length of $N_e$. 
Commands containing fewer than $N_e$ (sub-)tokens are zero-padded, whereas longer commands are truncated to match the predefined embedding dimension.
Note that the proposed formulation is generally agnostic to sequence length and can be adapted to variable lengths by using End-of-Sentence (EOS) tokens. 

\begin{figure}[t]
    \centering
    \includegraphics[width=0.95\linewidth]{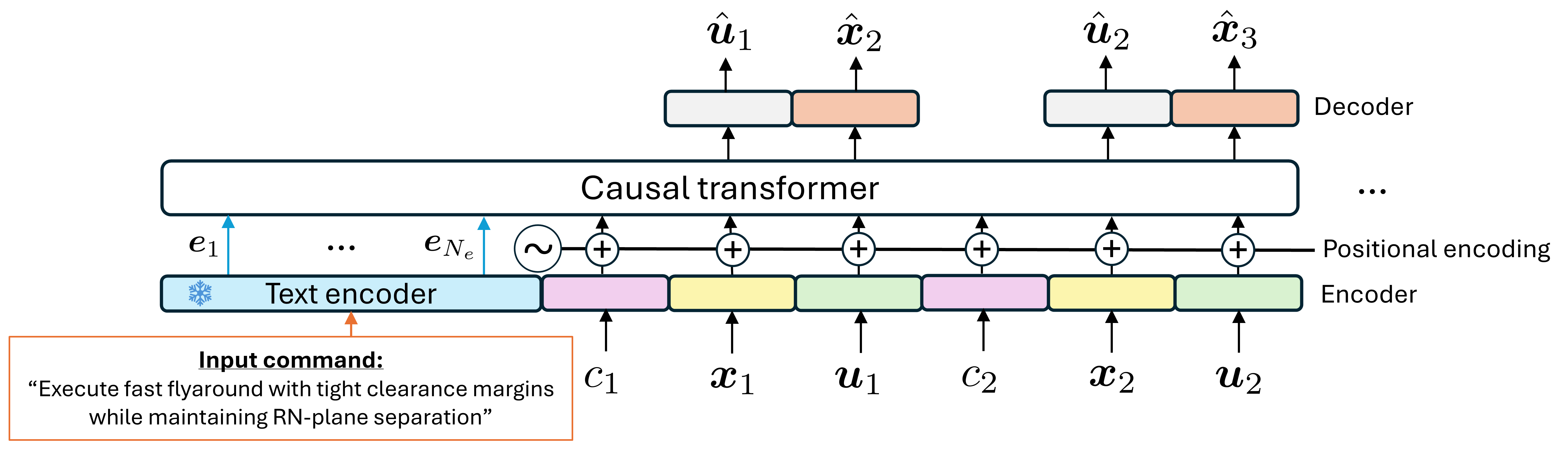}
    \caption{\edit{Causal transformer architecture encoding text commands and trajectory tokens (constraints, states, controls) into a shared latent space for autoregressive trajectory generation.}}
    \label{fig:tc_dt}
\end{figure}

To process a sequence of tokens in Eq.~\eqref{eq:traj_token_sages} and generate trajectories, a causal transformer that generates trajectories in the joint space of textual information and constraint signals is introduced.
\edit{As illustrated in Fig.~\ref{fig:tc_dt}, the input command is first encoded into a sequence of text embeddings, which serve as a prefix to the trajectory tokens. 
The trajectory tokenized sequence of constraint, state, and control tokens $(c_t, \boldsymbol{x}_t, \boldsymbol{u}_t)$ is all embedded into a shared latent space with positional encoding. 
The causal transformer processes this sequence autoregressively and predicts the next-step state and control $(\hat{x}_{t+1}, \hat{u}_{t+1})$ conditioned on all preceding tokens.}

\subsection{\edit{Workflow}}

The workflow of SAGES comprises four phases: (i) dataset generation, (ii) transformer training, (iii) test-time inference, and (iv) trajectory refinement via SCP. 

\subsubsection{\edit{Dataset Generation}}
A paired dataset of trajectories and high-level text commands must first be constructed to train the model. 
Each dataset element consists of a spacecraft trajectory and a natural-language description of the associated behavior.
The goal is to achieve broad diversity across operating conditions, problem parameters, and linguistic expressions.
This diversity is critical for the learning model to learn across a wide solution space and generalize in both control performance and semantic interpretation.
In this work, a scalable language-annotation pipeline for spacecraft trajectories enabled by LLMs \edit{is proposed}. 
A finite set of high-level behavior modes is first defined, each characterized by an ordered sequence of waypoints and the corresponding times at which they are reached. 
For each behavior mode, a large pool of trajectories satisfying these waypoint constraints is generated using a nonconvex optimization framework (SCP). 
In parallel, a separate and typically smaller pool of text commands is produced using GPT-4o API \cite{openai_gpt4o_2024}, which greatly accelerates the generation of diverse natural-language descriptions.
Because the trajectories and commands are created independently, their pool sizes may differ. 
After generation, the two pools are randomly shuffled, and command-trajectory pairs associated with the same behavior mode are drawn to form the final dataset. 
This decoupled procedure allows large-scale trajectory synthesis while keeping the language-generation burden manageable.
The detailed dataset-generation procedure adopted in this paper is provided in \edit{Appendix A}.

\subsubsection{\edit{Training}}

\edit{The transformer is trained via supervised behavior cloning in an autoregressive manner, where the model predicts the next state and control input conditioned on the ground-truth state–action history \cite{chen2021decision}. }
Given an input command describing the high-level intent of the maneuver, the text encoder produces a semantic embedding (cf. Eq.~\eqref{eq:txt_encoder}). 
Because the command corresponds to a trajectory-level instruction rather than a token-ordered sequence, no positional encoding is applied to the text embedding \cite{putterman2021pretraining}.
This embedding is then concatenated with the constraint signal and the state–action history and fed into a causal transformer. 
The transformer’s output layer splits into two task-specific heads that predict the next state and the next control input, respectively, thereby enabling joint autoregressive rollout conditioned on both semantics and dynamics.
\edit{The training objective minimizes the mean squared error between predicted and ground-truth trajectories, which is computed over a batch of $B$ trajectories as:}
\small
\begin{align}
    \mathcal{L}(\boldsymbol{\tau}_{1:N}) = \sum_{b=1}^{B} \sum_{k=1}^{N} \left( \left\| \boldsymbol{x}_k^{(b)} - \hat{\boldsymbol{x}}_k^{(b)} \right\|_2^2 + \left\| \boldsymbol{u}_k^{(b)} - \hat{\boldsymbol{u}}_k^{(b)} \right\|_2^2 \right),
\end{align}
\normalsize
where $\hat{\boldsymbol{x}}_k$ and $\hat{\boldsymbol{u}}_k$ denote the model’s predicted state and control input, and $\boldsymbol{x}_k$ and $\boldsymbol{u}_k$ are their corresponding ground-truth values sampled from the dataset.

\subsubsection{\edit{Test-time inference}}
After training, the transformer can be used to autoregressively generate spacecraft trajectories for novel input commands.
During the test-time inference, only the control input prediction head is utilized to generate the entire trajectory in an autoregressive manner.
At each step, the predicted control input is propagated through a known dynamics model $\boldsymbol{F}_k(\boldsymbol{x}_k, \boldsymbol{u}_k)$, which is the same model employed within the subsequent trajectory optimization problem.
This model-in-the-loop rollout ensures that the generated trajectories remain dynamically feasible.
The constraint-to-go is updated at every step based on the applied control input and the propagated state.

\subsubsection{\edit{Trajectory Refinement via SCP}}
The generated trajectories are then used to provide a warm-start for the SCP. 
Therefore, the SCP is used to map the initial guess onto the feasible domain.
In particular, SAGES solves the following problem as a particular form of the feasibility problem in Eq.~\eqref{eq:feas_prob}:
\begin{subequations}
\begin{align} \label{eq:obj_feasibility}
\min_{\{\boldsymbol{x}_k\}_{k=1}^{N}, \{\boldsymbol{u}_k\}_{k=1}^{N}} \ &  \tilde{\mathcal{J}} = \mathcal{J} + \sum_{k=1}^{N} \lambda  \left\| \boldsymbol{x}_k - \hat{\boldsymbol{x}}_k \right\|_2^2
 \\
\text{subject to} \quad & \eqref{eq:ocp_con_dyn}, \eqref{eq:ocp_con_path}
\end{align}
\end{subequations}
This choice of the objective function ensures that the refined trajectory remains close to the warm-start solution, thereby preserving its semantic behavior.
Here, $\lambda > 0$ is a hyperparameter that controls the trade-off between adherence to the generated trajectory and performance under the original objective. 
This approach contrasts with naive projection onto the feasible domain, such as in Algorithm 1 in \cite{mao2021scvx-fast}, as well as with traditional objectives such as fuel minimization. 
Such alternatives may yield feasible trajectories that are semantically inconsistent with the original intent, since the solution can deviate substantially from the warm-start.
Increasing $\lambda$ enforces closer adherence to the generated trajectory at the expense of higher control effort in the original objective. 
The appropriate value of $\lambda$ is scenario dependent and is ultimately selected by the designer to balance this trade-off.
The main advantage of this framework is that, although SAGES can generate semantically correct and safe trajectories, the refinement stage always prioritizes the feasibility of the trajectory. 
This hierarchy, where safety is prioritized over semantic alignment, is essential in safety-critical applications such as on-orbit maneuvering, where constraint violations cannot be tolerated.

\section{Problem Scenarios} \label{sec:case_study}

Two case studies are conducted to evaluate the performance of the proposed semantic trajectory generation framework: a free-flyer robotic testbed and a fault-tolerant spacecraft proximity operation scenario.
This section describes the problem formulation for each scenario and the detailed procedure of dataset generation.

\subsection{Free-flyer Robotic Testbed}

\begin{figure}[t]
    \centering
    \begin{subfigure}{0.44\linewidth}
        \centering
        \includegraphics[width=\linewidth]{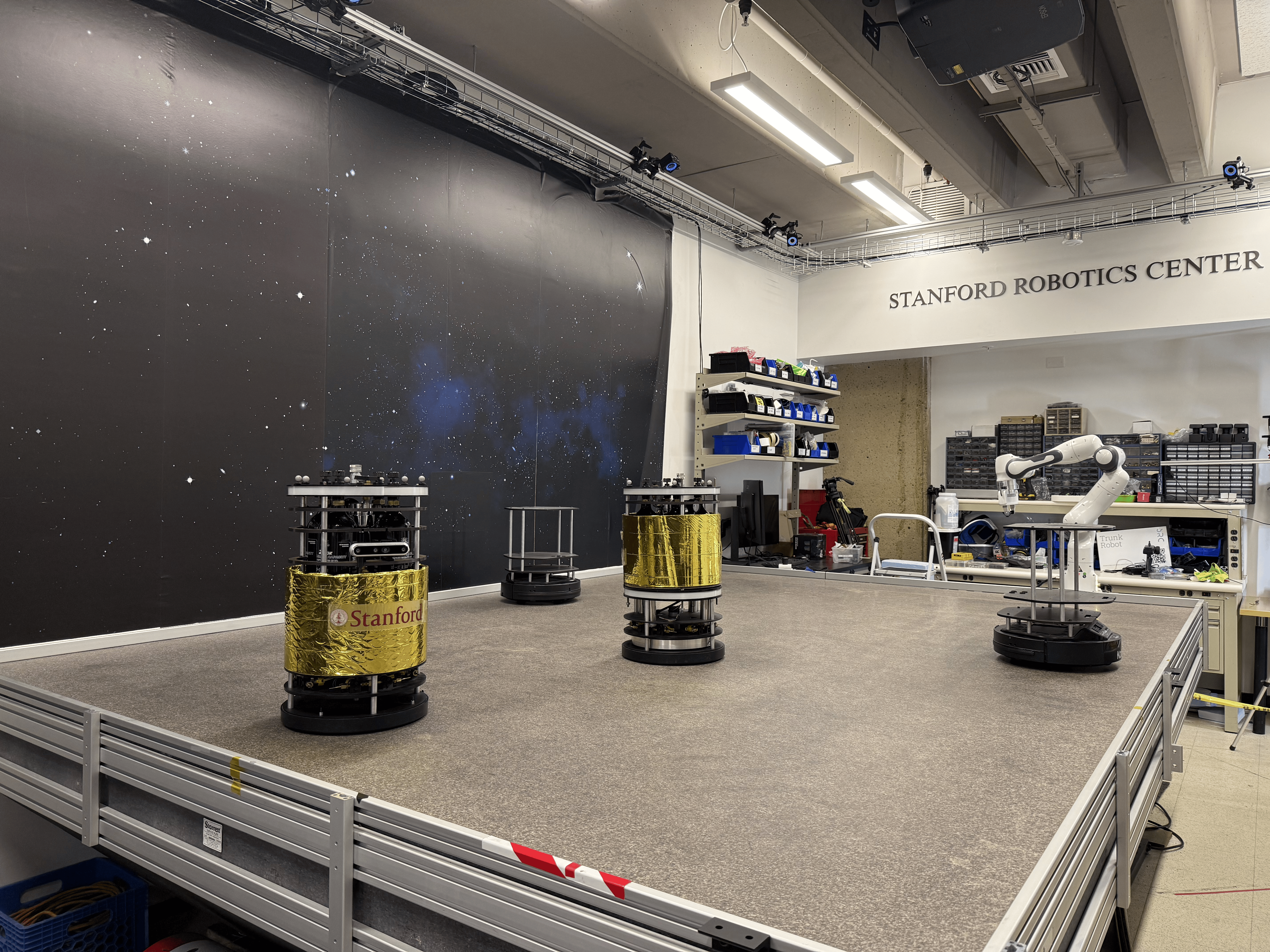}
        \caption{Hardware testbed on a quasi-frictionless table.}
        \label{fig:ff_sub1}
    \end{subfigure}
    \hfill
    \begin{subfigure}{0.55\linewidth}
        \centering
        \includegraphics[width=\linewidth]{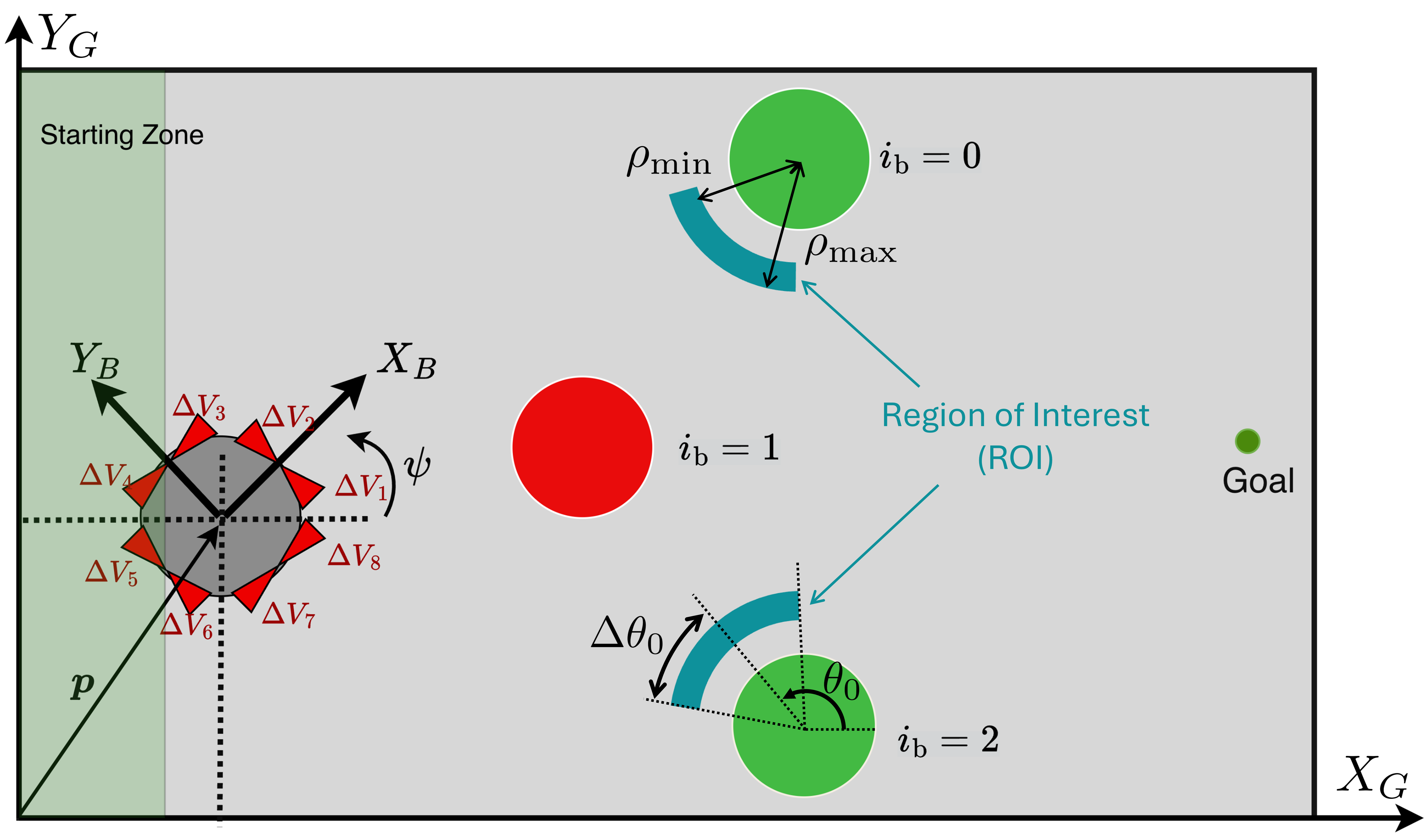}
        \caption{Reference frames, thruster configurations, and obstacle locations.}
        \label{fig:ff_sub2}
    \end{subfigure}
    \caption{Overview of the free-flyer robotic testbed.}
    \label{fig:free_flyer}
\end{figure}

The free-flyer system is a planar, microgravity robotic testbed that provides two translational and one rotational degree of freedom on a quasi-frictionless granite table. 
It is widely used as a ground-based platform for validating control algorithms for space applications.
\edit{Figure~\ref{fig:ff_sub1} shows the physical hardware, whereas Fig.~\ref{fig:ff_sub2} illustrates the coordinate systems that are used in the following trajectory optimization: the global Cartesian reference frame $\mathcal{O}_{G}$ and the free-flyer's body-fixed reference frame $\mathcal{O}_{B}$.}
The vehicle is actuated by eight on--off compressed $\text{CO}_2$ thrusters that generate an impulsive velocity change $\Delta V$ to achieve roto-translational motion. 
The workspace contains a central obstacle \edit{($i_b = 1$, red in Fig.~\ref{fig:ff_sub2})} and two \edit{observation targets ($i_b = 0$ and $2$, green in Fig.~\ref{fig:ff_sub2})} placed on the left and right sides of the table. 
The free-flyer initiates its maneuver from a predefined start region and approaches a fixed goal state, with these bodies creating two flanking corridors to the goal.
The state of the free-flyer is represented as $\boldsymbol{x} := [\boldsymbol{p}, \boldsymbol{v}, \psi, \omega] \in \mathbb{R}^6$, where $\boldsymbol{p}, \boldsymbol{v} \in \mathbb{R}^2$ are the two-dimensional position and velocity of its center of mass in the global frame, and $\psi, \omega \in \mathbb{R}$ are the bearing angle from the $X_G$-axis and its angular velocity, respectively.

The following subsections outline the scenario design, covering (i) the definition of high-level behavior patterns, (ii) the nonconvex trajectory optimization used for dataset generation, and (iii) the subsequent nonconvex refinement stage.

\subsubsection{Behavior Patterns and Model Conditioning}\label{sec:ff_behav_pattern}

Table~\ref{tab:commands_ff} summarizes the six distinct behavior modes considered in this scenario and provides representative example commands for each.
These modes are defined based on (i) whether the vehicle approaches the goal via left-side passage, right-side passage, or a central traverse, and (ii) whether the traverse is fast or slow, which is dictated by the arrival time to the goal. 
In particular, the left and right passages are induced by requiring the trajectory to pass through a designated region of interest (ROI) \edit{near the observation targets} before reaching the goal.
\edit{The waypoint position within the ROI, denoted by $\boldsymbol{p}_{\mathrm{wyp}}$, is defined as:}
\begin{equation}
\boldsymbol{p}_{\mathrm{wyp}} = \boldsymbol{p}^{(i_b^*)} + \rho \begin{bmatrix}
\cos\theta \\ \sin\theta
\end{bmatrix}, 
\quad \theta \in [\theta_0\edit{^{(i_b)}}-\Delta \theta_0, \theta_0\edit{^{(i_b)}}+\Delta\theta_0], \ \rho\in [\rho_{\min}, \rho_{\max}], \ i_{b}^*
=
\begin{cases}
0, & b\in\{0,1\},\\[4pt]
2, & b\in\{2,3\},
\end{cases}
\label{eq:ff_wyp_sampled}
\end{equation}
\edit{
where $\boldsymbol{p}^{(i_b)} \in \mathbb{R}^2$ denotes the center of the $i_b$-th body.
Here, $\rho$ defines the radial offset from the target center and is chosen to exceed the combined radii of the free-flyer and the body with margin. 
The nominal bearing angle $\theta_0$ is defined with respect to the $X_G$-axis, and $\Delta\theta_0$ specifies the allowable angular deviation. 
The parameters $(\rho,\theta)$ are sampled uniformly within their respective bounds. 
A schematic of the ROI geometry is shown in Fig.~\ref{fig:ff_sub2}.}
During annotation with text commands (cf. Table~\ref{tab:commands_ff}), each sentence is restricted to have fewer than 15 words, whereas the maximum text-embedding length is set to $N_e = 30$. 
\begin{table}[t]  \footnotesize
    \centering
    \caption{\edit{Behavior modes and sample text commands defined in the free-flyer scenario.}}
    \renewcommand*{\arraystretch}{1}
    \begin{tabular}{ll}
    \thickhline
        Behavior & Example Command \\
        \hline 
         0: Left passage (fast) & \makecell[l]{"Execute a rapid left-side bypass, maintaining tight KOZ clearance with an agile maneuver profile."} \\
         1: Left passage (slow) &  \makecell[l]{"Adopt a broad left-side arc, ensuring KOZ compliance with extended loiter in the central corridor."} \\
         2: Right passage (fast)  & \makecell[l]{"Clearance maintained; sharp right-side arc executed with compressed schedule, ensuring KOZ compliance."}\\
         3: Right passage (slow)  & \makecell[l]{"The trajectory follows a broad right-side arc, widening clearance for extended low-\(\Delta V\) standoff."} \\
         4: Central traverse (fast)  & \makecell[l]{"With precise KOZ compliance, a rapid central corridor sprint is executed via high RCS cadence."} \\
         5: Central traverse (slow) & \makecell[l]{"Prioritize broad central corridor transit, ensuring KOZ compliance with minimal \(\Delta V\) and cautious velocity."}\\
    \thickhline
    \end{tabular}
    \label{tab:commands_ff}
\end{table}

Based on the definition of the behavior modes, the semantic correctness is assessed by the following two criteria: 
\begin{enumerate}
    \item \edit{ROI passage:} The trajectory must pass through the designated ROI if specified (i.e., left and right passages). 
    \item \edit{Terminal-time consistency:} \edit{For fast traversal modes, the free-flyer must reach the goal position within an error of $\epsilon_g$ m by $t_f^{\mathrm{fast}} \leq 30~\mathrm{s}$. For slow traversal modes, the terminal time is required to lie within $t_f \in  [35, 40]~\mathrm{s}$.}
\end{enumerate}
These criteria ensure alignment between high-level natural-language intent and the generated motion profile to evaluate the semantic correctness of the generated trajectories.

\subsubsection{Trajectory Optimization for Dataset Generation}
The trajectory optimization problem of the free-flyer system used for the dataset generation is formulated as follows:
\begin{subequations} \label{eq:ocp_ff}
\begin{align}
    \min_{\{\boldsymbol{x}_k\}_{k=1}^{N}, \{\boldsymbol{u}_k\}_{k=1}^{N}} \quad & 
    \sum_{k=1}^{N} \lVert \boldsymbol{u}_k \rVert_1
    + w \sum_{k \in \mathcal{I}_{\text{wyp}}}
        \left(\lVert \boldsymbol{p}_k - \boldsymbol{p}_{\text{wyp}} \rVert_2 - r_{\text{wyp}}\right)^2, \label{eq:ocp_ff_cost} \\
    \text{subject to} \quad 
    & \boldsymbol{x}_{k+1} = \boldsymbol{x}_k + [\boldsymbol{I}_3 \cdot \Delta t , \boldsymbol{I}_3]^\top \boldsymbol{u}_k, 
    && \forall k = 1,...,N-1, \label{eq:ocp_ff_dyn} \\
    & \boldsymbol{x}_1 = \boldsymbol{x}_{i}, \quad
      \boldsymbol{x}_{N} = \boldsymbol{x}_{f}, \label{eq:ocp_ff_bc} \\
    & \boldsymbol{x}_k \in \mathcal{X}_{\text{table}}, 
    && \forall k = 1,...,N, \label{eq:ocp_ff_table} \\
    & \lVert \boldsymbol{p}_{k_{\text{wyp}}} - \boldsymbol{p}_{\text{wyp}} \rVert_2 \;\le\; r_{\text{wyp}}, \label{eq:ocp_ff_wyp}\\
    & 0 \le \Lambda^{-1} \mathbf{R}_{GB}^{-1}(\psi_k) \boldsymbol{u}_k \le \Delta V_{\max}, 
    && \forall k = 1,...,N, \label{eq:ocp_ff_thrust} \\
    & \bigl\| \boldsymbol{p}_k - \boldsymbol{p}_O^{(i_b)} \bigr\|_2 - \edit{R} \ge 0, &&  \forall k = 1,...,N,\ \forall i_b \in \{0,1,2\}.  \label{eq:ocp_ff_obs} 
\end{align}
\end{subequations}
First, Eq.~\eqref{eq:ocp_ff_cost} specifies the objective based on the impulsive-thruster control.
Eqs.~\eqref{eq:ocp_ff_dyn} and \eqref{eq:ocp_ff_bc} define the system dynamics and boundary conditions, and the testbed workspace limits are specified in Eq.~\eqref{eq:ocp_ff_table}. 
For behavior modes involving left- or right-passage, waypoint passage is enforced through two complementary mechanisms. 
First, Eq.~\eqref{eq:ocp_ff_wyp} imposes a hard, pointwise waypoint constraint, requiring the free-flyer’s position to lie within a radius $r_{\text{wyp}}$ of the sampled waypoint location (cf.~Eq.~\eqref{eq:ff_wyp_sampled}) at timestep $k_{\text{wyp}}$. 
Second, Eq.~\eqref{eq:ocp_ff_cost} includes a soft waypoint-loitering penalty, defined as the cumulative squared displacement from the waypoint over $\mathcal{I}_{\text{wyp}} = \{k_{\text{wyp}}-3,\dots,k_{\text{wyp}}+3\} \cap \{1,\dots,N\},$ and weight $w=10^3$. 
Whereas the hard constraint enforces passage at a specific time, the soft penalty encourages the vehicle to remain near the waypoint over a short interval surrounding the passage, facilitating the smooth maneuver near the ROI. 
Furthermore, Eq.~\eqref{eq:ocp_ff_thrust} denotes the nonconvex control input constraint.
The control input vector in the global frame is given by 
$\boldsymbol{u} = \mathbf{R}_{GB}(\psi)\, \Lambda \, \Delta \boldsymbol{V} \in \mathbb{R}^3,$ where $\Delta \boldsymbol{V} \in \mathbb{R}^8$ denotes the impulsive $\Delta V$s applied by each thruster; $\Lambda \in \mathbb{R}^{3\times 8}$ denotes the thruster configuration matrix; and $\mathbf{R}_{GB} \in \mathbb{R}^{3\times 3}$ denotes the rotation matrix from the body-fixed frame to the global frame.  
The $\Delta V$s applied by each thruster are physically limited to $0 \leq \Delta V \leq \Delta V_{\max} = {T \Delta t}/{m}$, where $T$ and $m$ denote the maximum thrust level and the system mass, respectively.
Finally, Eq.~\eqref{eq:ocp_ff_obs} represents the nonconvex collision avoidance constraint with respect to the three obstacles, \edit{where the distance between the centers of the obstacles and FF is larger than the sum of their radii, denoted as $R$}.
Appendix B presents the problem-specific parameters used in the free-flyer scenario. 
To generate "fast" behaviors, the terminal time is shortened from the nominal \(t_f = 40\,\mathrm{s}\) to \edit{
\(t_f^{\mathrm{fast}} \in [26,30] \) s, where the problem is solved under a reduced horizon $N_{\mathrm{fast}} = \lfloor t_f^{\mathrm{fast}} / \Delta t \rfloor$.}

The trajectory dataset of the free-flyer scenario is generated by successively solving SCPs.
First, given an initial state, transfer time, and waypoint condition (position and timestep), the SCP solves a relaxed nonconvex problem in Eq.~\eqref{eq:ocp_ff} without Eq.~\eqref{eq:ocp_ff_obs}, which generates a dynamically feasible trajectory that satisfies the waypoint passage but not the collision avoidance constraints.
The resulting trajectory is then used as a warm-start for the second SCP that solves the full problem in Eq.~\eqref{eq:ocp_ff}. 
This two-stage SCP structure reduces the infeasibility in the SCP and ensures that the final trajectories satisfy both the behavior-conditioned waypoint and collision avoidance. 

\begin{figure}[t]
    \centering
    \includegraphics[width=0.95\linewidth]{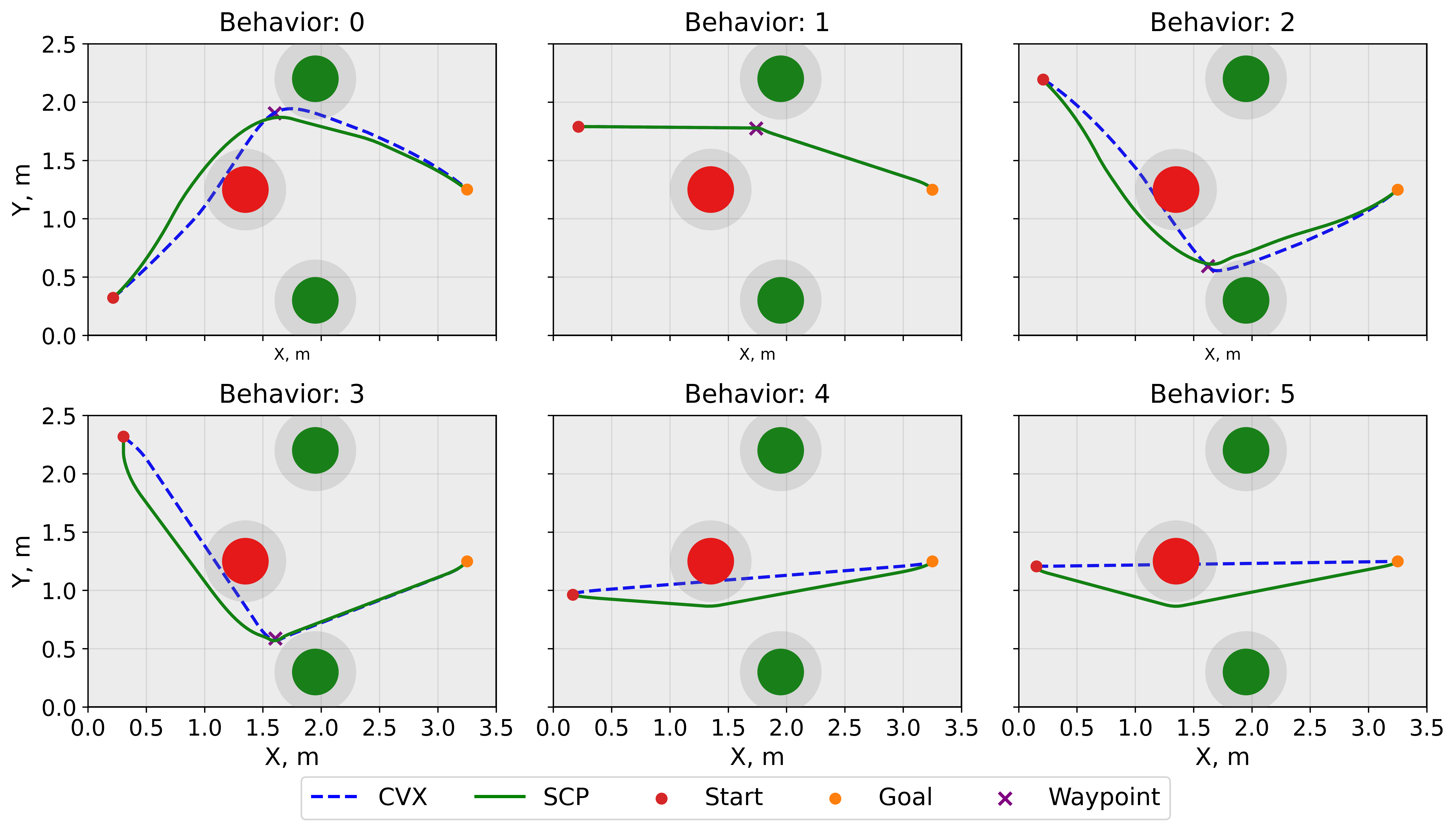}
    \caption{\edit{Representative free-flyer trajectories corresponding to behavior modes defined in Table~\ref{tab:commands_ff}}.}
    \label{fig:ff_behav}
\end{figure}

Based on the above formulation, sample trajectories demonstrating the behavior patterns defined in Table~\ref{tab:commands_ff} are shown in Fig.~\ref{fig:ff_behav}.

\subsubsection{Trajectory Optimization for Refinement after the Test-time Inference}

After the warm-start is generated, the following optimization problem is solved for the trajectory refinement:
\begin{subequations} \label{eq:ocp_ff_feasibility}
\begin{align}
\min_{\{\boldsymbol{x}_k\}_{k=1}^{N}, \{\boldsymbol{u}_k\}_{k=1}^{N}} \quad & \eqref{eq:obj_feasibility} \quad  
\text{subject to} \quad 
\eqref{eq:ocp_ff_dyn}, \eqref{eq:ocp_ff_bc}, \eqref{eq:ocp_ff_table}, \eqref{eq:ocp_ff_thrust}, \eqref{eq:ocp_ff_obs}.
\end{align}
\end{subequations}
\edit{Note that no waypoint-passage constraint is imposed during refinement, since explicit waypoint information is assumed to be unavailable at deployment. Instead, such waypoint structure is expected to be implicitly encoded in the generated trajectory.}

\subsection{Spacecraft Proximity Operation}

The second case study considered in this paper is the spacecraft proximity operation scenario, in which a servicer spacecraft executes a complex maneuver around a target spacecraft.
Due to the inherent difficulty of this maneuver, a conservative continuous-time passive safety constraint is imposed in the trajectory optimization, making strict constraint satisfaction particularly challenging.
Additionally, the high-level text commands specify the mission-related parameters, such as time durations of the sub-tasks within each maneuver, using explicit numerical values. 
Although this mixture of natural language and numerical information reflects realistic spacecraft operations, it introduces additional learning challenges compared to the free-flyer scenario. 
In particular, the model must accurately capture the temporal structure and coherence of each behavior, while also extracting the numerical quantities embedded in the text and associating them with the corresponding maneuver pattern.

The state vector $\boldsymbol{x} \in \mathbb{R}^6$ is represented in the form of Relative Orbital Elements (ROE)~\cite{damico_phd_2010}.  
Let the Keplerian elements be denoted by $\boldsymbol{\alpha} = [a, e, i, \Omega, \omega, M]$, where $a$ is the semi-major axis, $e$ the eccentricity, $i$ the inclination, $\Omega$ the right ascension of the ascending node, $\omega$ the argument of periapsis, and $M$ the mean anomaly.  
The ROE are defined as specific differences between the orbital elements of the servicer (deputy) and the target, denoted by $\boldsymbol{\alpha}_d$ and $\boldsymbol{\alpha}$, respectively.  
In this study, the quasi-nonsingular ROE (qnsROE) formulation scaled by the chief's semimajor axis is adopted, expressed as~\cite{damico_phd_2010}:
\begin{equation}
\boldsymbol{x}_{\text{qns}} = a
\begin{bmatrix}
\delta a \\ \delta \lambda \\ \delta e_x \\ \delta e_y \\ \delta i_x \\ \delta i_y
\end{bmatrix}
= a
\begin{bmatrix}
(a_d - a)/a \\
(M_d + \omega_d) - (M + \omega) + (\Omega_d - \Omega)\cos i \\
e_d\cos\omega_d - e\cos\omega \\
e_d\sin\omega_d - e\sin\omega \\
i_d - i \\
(\Omega_d - \Omega)\sin i
\end{bmatrix}.
\end{equation}
\edit{One advantage of using the qnsROE is in the geometric relationship for the construction of safe relative orbits, known as E/I-vector separation \cite{damico_proximity_2006}. In this framework, the relative motion is decomposed into relative eccentricity vector $\delta \boldsymbol{e} = [\delta e_x, \delta e_y]$ and inclination vector $\delta \boldsymbol{i} = [\delta i_x, \delta i_y]$, and safety is achieved by designing their relative orientation such that the resulting trajectory avoids penetration of the KOZ in the Radial-Normal (RN) plane.}

In parallel with the previous section, the subsequent subsections describe the scenario design, covering (i) the definition of high-level behavior patterns, (ii) the nonconvex trajectory optimization used for dataset generation, and (iii) the feasibility problem employed for final nonconvex refinement.

\subsubsection{Behavior Patterns and Model Conditioning}\label{sec:rpod_behav_pattern}

Six behavior modes are considered to describe the maneuver of the servicer. 
The servicer spacecraft begins its maneuver from an initial along-track offset of approximately $\delta\lambda = -120 \ \text{m}$ (the -V-bar direction) with a relative orbit state that exhibits no RN-plane separation, \edit{i.e., $\boldsymbol{x}_{\mathrm{qns}}(t_0) = [0, -120, 0,0,0,0]$ m}. 
From this \edit{initial} configuration, the servicer performs various behaviors such as circumnavigation or docking.
Furthermore, this case study does not involve conditioning based on the goal state; namely, the trained model is desired to define its (terminal) waypoints \edit{based on the trajectory} rollout.

\edit{
To realize these behaviors within the dataset, waypoints are strategically selected, as summarized in Table~\ref{tab:wyp_rpod}, which specifies the admissible waypoint-passage timesteps and the corresponding ROE waypoint.
During dataset generation, the waypoint-passage timestep is randomly sampled as an integer from the specified range, while the state is sampled from the discretized grid. Specifically, the notation \( z = x \pm \Delta x \) denotes that \(z\) denotes a uniform sampling from \(x - \Delta x \le z \le x + \Delta x\), discretized into 21 values.
Note that the perturbation in $\delta a$ is set to zero to ensure bounded relative motion at each waypoint. 
Furthermore, as the dominant perturbation mode in the ROE space arises from the drift in $\delta \lambda$ induced by nonzero $\delta a$ \cite{damico_phd_2010}, the randomization applied to $\delta \lambda$ is assigned a larger variance than that of the relative eccentricity $\delta \boldsymbol{e}$ and inclination $\delta \boldsymbol{i}$. 
The nominal waypoint configuration enforces parallel $\delta \boldsymbol{i}$ and $\delta \boldsymbol{e}$ with $\delta i_x = 0$, thereby maximizing the planar separation in the RN plane while suppressing the secular drift induced by the $J_2$ perturbation \cite{damico_phd_2010}. 
The waypoint ranges are manually tuned to maintain a roughly uniform dataset across behavior modes, accounting for the fact that SCP does not yield feasible trajectories for all sampled cases.}

\edit{The notions of fast and slow maneuvers are encoded explicitly through numerical values in the input commands.
As each command may specify a sequence of actions, interpreting its semantics and executing a safe maneuver becomes more challenging. 
Furthermore, the inclusion of numerical values in the command sentences introduces additional complexity in the dataset annotation process.}
To efficiently characterize trajectories and label each with an appropriate quantitative description, a collection of pre-generated command templates is first prepared, each capable of accepting a set of behavior-dependent variables. 
After generating the dataset trajectories, a template corresponding to the desired behavior mode is randomly drawn from this pool, and quantitative values, such as the terminal along-track separation ($a\delta\lambda$) or waypoint epochs, are substituted to form a trajectory-specific command.
\edit{Since the numerical values are instantiated from the specific trajectory, each resulting command becomes trajectory-specific.}
The variables associated with each behavior mode, along with illustrative command templates, are summarized in Table~\ref{tab:commands_rpod}. 
\edit{Although a command may describe a sequence of actions, the semantic-correctness test is performed with respect to the instantiated intermediate waypoint condition and the terminal state associated with the selected behavior mode.}
To promote linguistic diversity, a given command is not required to include all available variables.
During the auto-generation of input commands, each sentence is set to have fewer than 23 words, and the maximum text-embedding length is set to $N_e = 50$. 
\begin{table}[t]\footnotesize
    \caption{\edit{Definition of waypoints for each behavior mode in the spacecraft proximity operation scenario. }}
    \centering
    \begin{tabular}{l cc}
    \thickhline
        \multirow{2}{*}{Behavior} & \multicolumn{2}{c}{Waypoints}  \\
        \cmidrule{2-3}
        & timestep, $k_{\text{wyp}}$ & $\boldsymbol{x}_{\text{qns}} \pm \Delta \boldsymbol{x}_{\text{qns}}  [\text{m}]$ \\
    \hline
    (initial state) & 
        $1$ & 
        $[0,\,-120\pm20,\;0\pm4,\;5\pm4,\;0\pm4,\;5\pm4]$ \\
    \hline
    0: Approach to the target and circumnavigate &
        \makecell[c]{
            $[31,49]$ \\ $50$
        } &
        \makecell[c]{
            $[0,\;0\pm5,\;0\pm2,\;32\pm2,\;0\pm2,\;32\pm2]$ \\
            $[0,\;0\pm5,\;5\pm2,\;32\pm2,\;5\pm2,\;32\pm2]$
        } \\
    \hline
    1: Go to -35 m V-bar and hold (dock) &
        \makecell[c]{
            $[31,49]$ \\
            $50$
        } &
        \makecell[c]{
            $[0,\;-35\pm5,\;0\pm2,\;0\pm2,\;0\pm2,\;0\pm2]$ \\
            $[0,\;-35\pm5,\;0\pm2,\;0\pm2,\;0\pm2,\;0\pm2]$
        } \\
    \hline
    2: Fast flyby under KOZ, from -V-bar to +V-bar &
        \makecell[c]{
            $[35, 49]$ \\ $50$
        } &
        \makecell[c]{
            $[0,\;150\pm2,\;0\pm2,\;5\pm2,\;0\pm2,\;5\pm2]$ \\
            $[0,\;150\pm2,\;0\pm2,\;5\pm\edit{2},\;0\pm2,\;5\pm2]$
        } \\
    \hline
    3: Flyby with RN (Radial-Normal)-plane separation, from -V-bar to +V-bar &
        \makecell[c]{
            $[6,15]$ \\ $[36,45]$\\ $50$
        } &
        \makecell[c]{
            $[0,\;-120\pm20,\;0\pm2,\;25\pm2,\;0\pm2,\;25\pm2]$ \\
            $[0,\;120\pm20,\;0,\;25\pm2,\;0,\;25\pm2]$ \\
            $[0,\;120\pm10,\;0,\;25\pm2,\;0,\;25\pm2]$
        } \\
    \hline
    4: Approach, circumnavigate, then escape along +V-bar &
        \makecell[c]{
            $[21,25]$ \\ $[35,39]$ \\ $50$ 
        } &
        \makecell[c]{
            $[0,\;0,\;0,\;32\pm2,\;0,\;32\pm2]$ \\
            $[0,\;0,\;0,\;32\pm2,\;0,\;32\pm2]$ \\
            $[0,\;-120\pm10,\;0\pm2,\;35\pm2,\;0\pm2,\;35\pm2]$ 
        } \\
    \hline
    5: Approach, circumnavigate, then escape along -V-bar &
        \makecell[c]{
            $[21,25]$ \\ $[35,39]$ \\ $50$
        } &
        \makecell[c]{
            $[0,\;0,\;0,\;32\pm2,\;0,\;32\pm2]$ \\
            $[0,\;0,\;0,\;32\pm2,\;0,\;32\pm2]$ \\
            $[0,\;120\pm10,\;0\pm2,\;35\pm2,\;0\pm2,\;35\pm2]$ 
        } \\
    \thickhline
    \end{tabular}
    \label{tab:wyp_rpod}
\end{table}
\begin{table}[t]  \footnotesize
\caption{\edit{Behavior modes and sample command templates for the spacecraft proximity operation scenario.}}
    \centering
    \begin{tabular}{cll}
    \thickhline
        Behavior & \edit{Quantitative Parameters} & Example Command Template \\
    \hline
    0 & \edit{$T_{\text{appr}}$} &
    \makecell[l]{
        Execute spiral approach to target for \edit{$T_{\text{appr}}$} orbits; thereafter,\\
        circumnavigate while ensuring RN-plane separation.
    } \\
    1 & \edit{$T_{\text{appr}}$, $a\delta \lambda$} &
    \makecell[l]{
        Use \edit{$T_{\text{appr}}$} orbits to close to \edit{$a\delta \lambda$} m along -V-bar;\\
        remain there until instructed otherwise.
    } \\
    2 & \edit{$T_{\text{appr}}$, $a\delta \lambda$} &
    \makecell[l]{
        Execute a rapid underfly trajectory to +V-bar at \edit{$a\delta \lambda$} m in\\
        \edit{$T_{\text{appr}}$} orbits with focus on RT-plane safety.
    } \\
    3 & \edit{$T_{\text{EI}}$, $T_{\text{trans}}$} &
    \makecell[l]{
        Flyby operation; establish E/I-vector separation in \edit{$T_{\text{EI}}$} orbits; execute\\
        delta-a until \edit{$T_{\text{trans}}$} orbits; hold at -V-bar.
    } \\
    4 & \edit{$T_{\text{appr}}$, $T_{\text{circ}}$} &
    \makecell[l]{
        Commence -V-bar approach for \edit{$T_{\text{appr}}$} orbits; circumnavigate until\\
        \edit{$T_{\text{circ}}$} orbits; maneuver to +V-bar afterwards.
    } \\
    5 & \edit{$T_{\text{appr}}$, $T_{\text{circ}}$} &
    \makecell[l]{
        Initiate approach from -V-bar for \edit{$T_{\text{appr}}$} orbits; circumnavigate until\\
       \edit{$T_{\text{circ}}$} orbits; proceed back to -V-bar.
    } \\
    \thickhline
    \end{tabular}
    \label{tab:commands_rpod}
\end{table}

\edit{Semantic correctness is assessed by verifying whether a generated trajectory satisfies the prescribed waypoint conditions in Table~\ref{tab:wyp_rpod} after instantiating the quantitative parameters appearing in the command.}
\edit{Three quantities are evaluated:}
(i) the waypoint-passing epoch,
(ii) the commanded along-track separation, and
(iii) the terminal state.
For commands specifying a time index (e.g., $T_\text{appr}$), the value is converted to the closest discrete timestep, denoted by $k_{\mathrm{wyp}}$.
At this index, element-wise consistency with the waypoint across the \edit{six qnsROE components} is checked using:
\begin{align} \label{eq:rpod_wyp_dom}
\left|x_{k_{\mathrm{wyp}},j} - x_{\mathrm{wyp},j}\right|
\le q \left|\Delta x_j\right|, 
\qquad j = 1, \ldots, 6,
\end{align}
where $\boldsymbol{x}_{\mathrm{wyp}}$ is the nominal waypoint, and $\Delta \boldsymbol{x}$ denotes the vector of maximum admissible deviations used during waypoint generation, as summarized in Table~\ref{tab:wyp_rpod}. 
The hyperparameter $q>0$ adjusts the tolerance of the criterion; $q=1$ corresponds to the requirement that the generated trajectory falls within the same tolerance bounds used to construct the training dataset.
In Table~\ref{tab:wyp_rpod}, some waypoint components are strictly fixed, \edit{such as \(\delta a\). 
For these components, an additional margin of \(\Delta x_j = \pm 2\,\mathrm{m}\) is introduced in the waypoint-inclusion test.}
The same condition as in Eq.~\eqref{eq:rpod_wyp_dom} is applied to the terminal state to determine whether the trajectory converges to the designed terminal set, although this terminal requirement is not explicitly commanded in the text input.
\edit{Finally, for commands that specify an along-track separation \(a\delta\lambda\),} correctness is evaluated analogously:
\begin{align} \label{eq:rpod_wyp_dom_dlam} 
\color{black}
\bigl|a\delta\lambda_{k_{\mathrm{wyp}}} - a\delta\lambda_{\mathrm{wyp}}\bigr| \leq q\bigl|\Delta a\delta\lambda\bigr|,
\end{align}
where \edit{$a\delta\lambda_{\mathrm{wyp}}$ and $\Delta a\delta\lambda$} denote the nominal relative mean longitude and its allowable tolerance, respectively.

In summary, a trajectory is deemed semantically correct if both of the following conditions are satisfied:
\begin{enumerate}
    \item \edit{Terminal-state consistency}: the final qnsROE state lies within the admissible terminal waypoint set, i.e., it satisfies the tolerance condition in~\eqref{eq:rpod_wyp_dom} when evaluated at the terminal index.
    \item \edit{Waypoint or along-track consistency}: at the commanded epoch $k_{\mathrm{wyp}}$, the trajectory satisfies the corresponding inclusion condition for either (i) the prescribed waypoint or (ii) the commanded relative mean longitude $\delta\lambda$, depending on which quantity is explicitly referenced in the input command.
\end{enumerate}
These requirements ensure that semantic correctness accounts for both the long-horizon behavior of the optimized trajectory and the quantitative constraints specified in the text instruction.

\subsubsection{Trajectory Optimization for Dataset Generation}
To generate a set of trajectories that complies with the safety constraint, a discrete-time nonconvex optimal control problem for fault-tolerant spacecraft proximity operation is formulated as follows:
\begin{subequations}  \label{eq:ocp_rpod_nonconvex} 
\begin{alignat}{2}
    \min_{\{\boldsymbol{x}_k\}_{k=1}^{N}, \{\boldsymbol{u}_k\}_{k=1}^{N}} \ &  \sum_{k=1}^{N} \|\boldsymbol{u}_k\|_2, \label{eq:ocp_rpod_obj} \\
    \text{subject to} \quad & \boldsymbol{x}_{k+1} = \Phi(t_{k+1}, t_k)(\boldsymbol{x}_{k} + \Gamma_k \boldsymbol{u}_k), \quad && \forall \ k = 1,...,N-1,  \label{eq:ocp_rpod_dyn} \\
    & g(t_k;\tau,\alpha) := 1 - \boldsymbol{x}(t_k;\tau,\alpha)^\top S_{k\tau} \boldsymbol{x}(t_k;\tau,\alpha) \leq 0 \quad && \forall \ k = 1,...,N, \forall (\tau,\alpha) \in [0, \tau^s] \times [0,1] \label{eq:ocp_rpod_ps_ct} \\
    & \boldsymbol{x}_1 = \boldsymbol{x}_i, \quad  \boldsymbol{x}_{N} + \Gamma_N \boldsymbol{u}_N = \boldsymbol{x}_f, \label{eq:ocp_rpod_bc} \\
    & \color{black} \boldsymbol{x}_{{k}_{\text{wyp},l}} = \boldsymbol{x}_{\text{wyp},l}, \quad && \color{black} \forall \ {l} \in L \label{eq:ocp_rpod_wyp} \\ 
    \text{where} \quad 
    & \boldsymbol{x}(t_k; \tau, \alpha) = \boldsymbol{x}(t_k; 0, \alpha)  + \int_{0}^{\tau} f(\gamma, \boldsymbol{x}(t_k;\gamma), 0) d\gamma  = \; &&\Phi(t_k+\tau,t_k)\boldsymbol{x}(t_k; 0, \alpha), \\
    & \boldsymbol{x}(t_k; 0, \alpha) = \boldsymbol{x}_k + \alpha \Gamma_k \boldsymbol{u}_k, \\
    & S_{k\tau} = \Psi(t_k+\tau)^\top P \Psi(t_k+\tau). \label{eq:S_ktau}
\end{alignat}
\end{subequations}
The objective of the optimization problem is to minimize the total control cost, modeled as the sum of impulsive maneuver magnitudes.  
The discrete-time qnsROE dynamics, subject to impulsive velocity changes $\boldsymbol{u}_k \in \mathbb{R}^3$ applied in the Radial–Tangential–Normal (RTN) frame, are presented in Eq.~\eqref{eq:ocp_rpod_dyn}.  
\edit{The State Transition Matrix (STM) \( \Phi(t_{k+1}, t_k) \in \mathbb{R}^{6\times6} \), which incorporates the secular \(J_2\) perturbation \cite{koenig2017new}, and the control input matrix \( \Gamma_k = \Gamma(t_k) \in \mathbb{R}^{6\times3} \)  \cite{damico_phd_2010} are analytical functions of the target's orbital elements and are independent of the relative state itself. 
Accordingly, the dynamics can be expressed as a Linear Time-Varying (LTV) system.}
The boundary conditions and \edit{passage of $L$ waypoints} are imposed through Eqs.~\eqref{eq:ocp_rpod_bc} and ~\eqref{eq:ocp_rpod_wyp}, respectively. 

The most critical nonconvex constraint in this problem is the continuous-time passive-safety requirement under the contingency of partially imperfect burns, as expressed in Eq.~\eqref{eq:ocp_rpod_ps_ct}.  
This constraint ensures that the deputy remains outside the ellipsoidal keep-out zone (KOZ) surrounding the target continuously over the controlled interval $t \in [0, t_f]$.  
Safety is further enforced during a subsequent drift interval $\tau \in [0, \tau^s]$ following an unexpected loss of control authority at the post-burn state $\boldsymbol{x}(t_k;0,\alpha)$.
Passive safety must be guaranteed (i) continuously in time, rather than at discrete nodes, and (ii) even under incomplete burn execution, characterized by the burn completion factor $\alpha \in [0, 1]$.
Despite its conservatism, enforcing passive safety under imperfect burns is justified, as the objective is to enable complex maneuvers around the target where the geometric safety criteria \cite{damico_phd_2010} cannot always be guaranteed.
The geometry of each KOZ is specified by a diagonal matrix $P \in \mathbb{R}^{6\times6}$, which defines the ellipsoidal semi-axes in the RTN frame, and matrix $\Psi_k = \Psi(t_k) \in \mathbb{R}^{6\times6}$ denotes the first-order linear mapping from qnsROE to the relative Cartesian state in the RTN frame, denoted as $\boldsymbol{x}_{\text{RTN}}$~\cite{damico_phd_2010}. 
\edit{Therefore, \( S_{k\tau} \in \mathbb{R}^{6 \times 6} \) denotes the induced quadratic form in the ROE space that represents the KOZ constraint after propagation over the drift interval \( \tau \), as defined in Eq.~\eqref{eq:S_ktau}.}
Note that stochastic uncertainties due to actuation misalignment, navigation error, or unmodeled accelerations are not explicitly modeled in the chance-constrained framework; only the contingency arising from imperfect impulsive burns is considered in this problem.
This simplification is justified because close-proximity operations achieve high relative navigation accuracy \cite{Guffanti2023Visors}, and unmodeled accelerations are negligible over the few orbital revolutions considered.
Instead, emphasis is placed on ensuring continuous-time satisfaction of the passive safety constraint and robust exclusion from the KOZ under the presence of partially imperfect burns, which dominate the overall safety behavior.  

The continuous-time passive safety constraint for an impulsive control at $t_k$ is enforced by satisfying the following integral inequality~\cite{elango2025continuous}:
\begin{equation}\label{eq:ct_ps_integral}
g(t_k;\tau,\alpha) \leq 0,  \ \forall (\tau,\alpha) \in [0, \tau^s] \times [0,1] 
\quad \Leftrightarrow\quad 
\tilde{g}(x_k, u_k) := \int_{0}^{1}\int_{0}^{\tau^s} \left| g(t_k;\tau,\alpha) \right|_+^2 \, d\tau \ d\alpha \leq 0,
\end{equation}
where $|x|_{+} = \max(x,0)$ denotes the hinge function. 
The convexified constraint of Eq.~\eqref{eq:ct_ps_integral} using the reference variables $(\bar{x}_k, \bar{u}_k)$ is obtained as:
\begin{subequations}
\begin{align}
    & G_k^x (\boldsymbol{x}_k - \bar{\boldsymbol{x}}_k) + G_k^u (\boldsymbol{u}_k - \bar{\boldsymbol{u}}_k) + \tilde{g}_k(\bar{\boldsymbol{x}}_k, \bar{\boldsymbol{u}}_k)= 0,  \label{eq:convexified_ps} \\
    & G_k^x := \left . \frac{\partial \tilde{g}_k(\boldsymbol{x}_k, \boldsymbol{u}_k)}{\partial \boldsymbol{x}_k} \right\vert_{(\bar{\boldsymbol{x}}_k, \bar{\boldsymbol{u}}_k)}
      = -4 \int_{0}^{1} \int_{0}^{\tau^s} \left| \bar{g}(t_k;\tau,\alpha) \right|_+ {\Phi}(t_k+\tau,t_k)^\top S_{k\tau}\, {\Phi}(t_k+\tau,t_k)\bar{\boldsymbol{x}}(t_k;0, \alpha)\; d\tau \ d\alpha \label{eq:Gk^x}, \\
    & G_k^u := \left . \frac{\partial \tilde{g}_k(\boldsymbol{x}_k, \boldsymbol{u}_k)}{\partial \boldsymbol{u}_k} \right\vert_{(\bar{\boldsymbol{x}}_k, \bar{\boldsymbol{u}}_k)}
      = -4 \int_{0}^{1} \int_{0}^{\tau^s} \left| \bar{g}(t_k;\tau,\alpha) \right|_+ \Gamma_k^\top {\Phi}(t_k+\tau,t_k)^\top S_{k\tau} \, {\Phi}(t_k+\tau;t_k)\bar{\boldsymbol{x}}(t_k;0, \alpha)\; d\tau \ d\alpha \label{eq:Gk^u},
\end{align}
\end{subequations}
where $\bar{g}(t_k; \tau,\alpha)$ is evaluated using the reference variables.  
While the original theoretical framework enforces continuous-time safety via Ordinary Differential Equation (ODE) integration to evaluate Eq.~\eqref{eq:convexified_ps} \cite{elango2025continuous}, this computation can be simplified for impulsive control under LTV dynamics. 
In particular, this paper replaces numerical propagation with a Gauss-Legendre quadrature rule to evaluate the constraint in Eq.~\eqref{eq:ocp_rpod_ps_ct} using the analytical STM.
The details of the constraint formulation and its convexification are provided in Appendix C. 
\edit{Figure~\ref{fig:ps_ct} illustrates an example trajectory with continuous-time passive safety with respect to the spherical KOZ with 25 m radius under partially imperfect burns.}
Particularly, Fig.~\ref{fig:ps_ct_range} illustrates the range between the servicer and the target, where the exclusion from the spherical KOZ is satisfied not only node-wise but also at the level of both nominal and drift continuous trajectories.
\edit{The color of the drift trajectories corresponds to the completion rate of the burn $\alpha \in [0,1]$.}
\begin{figure}[th!]
     \centering
     \begin{subfigure}[th!]{0.43\linewidth}
         \centering
         \includegraphics[width=\linewidth]{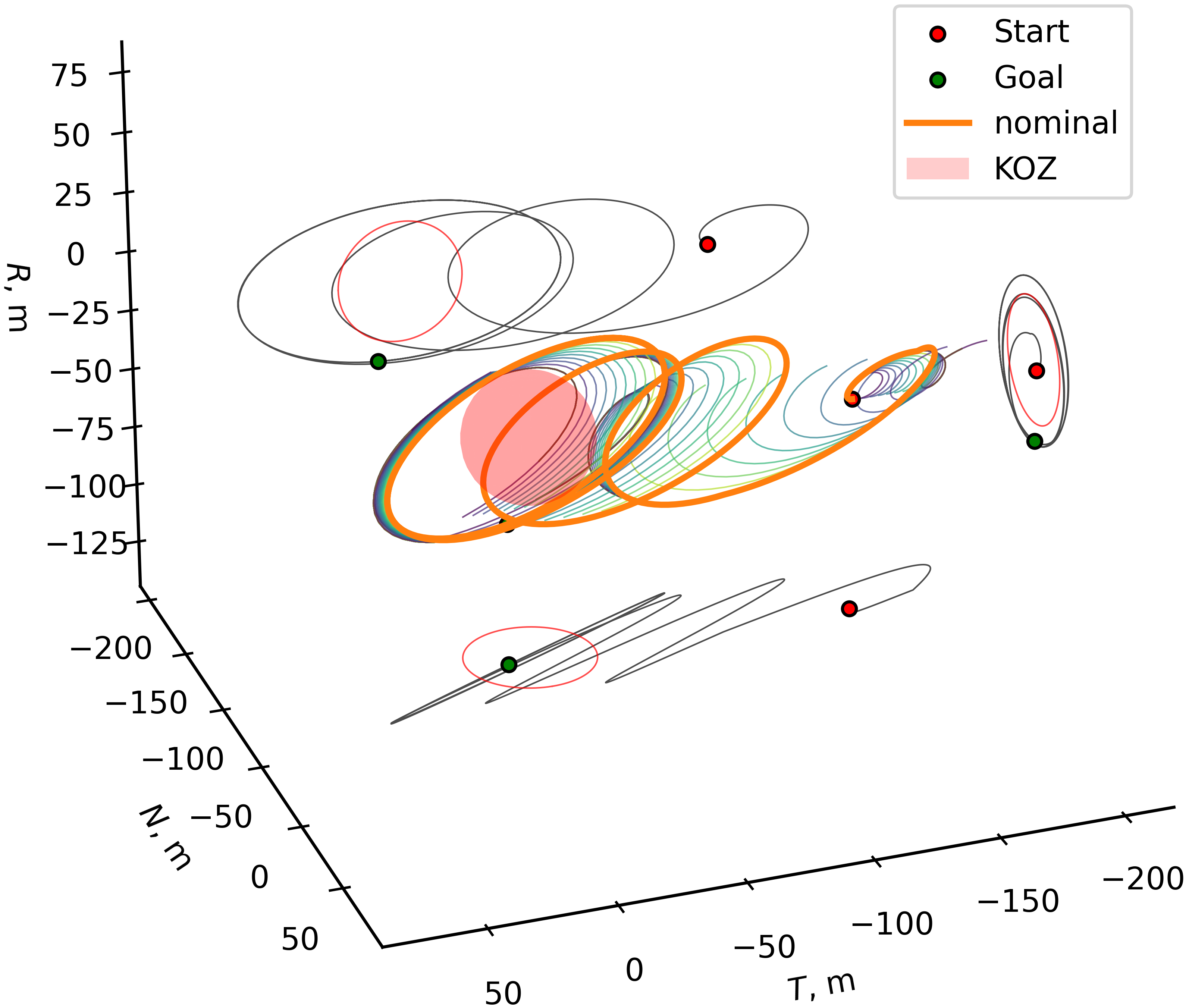 }
         \caption{RTN frame}
         \label{fig:ps_ct_traj}
    \end{subfigure} 
    \begin{subfigure}[th!]{0.56\linewidth}
         \centering
         \includegraphics[width=\linewidth]{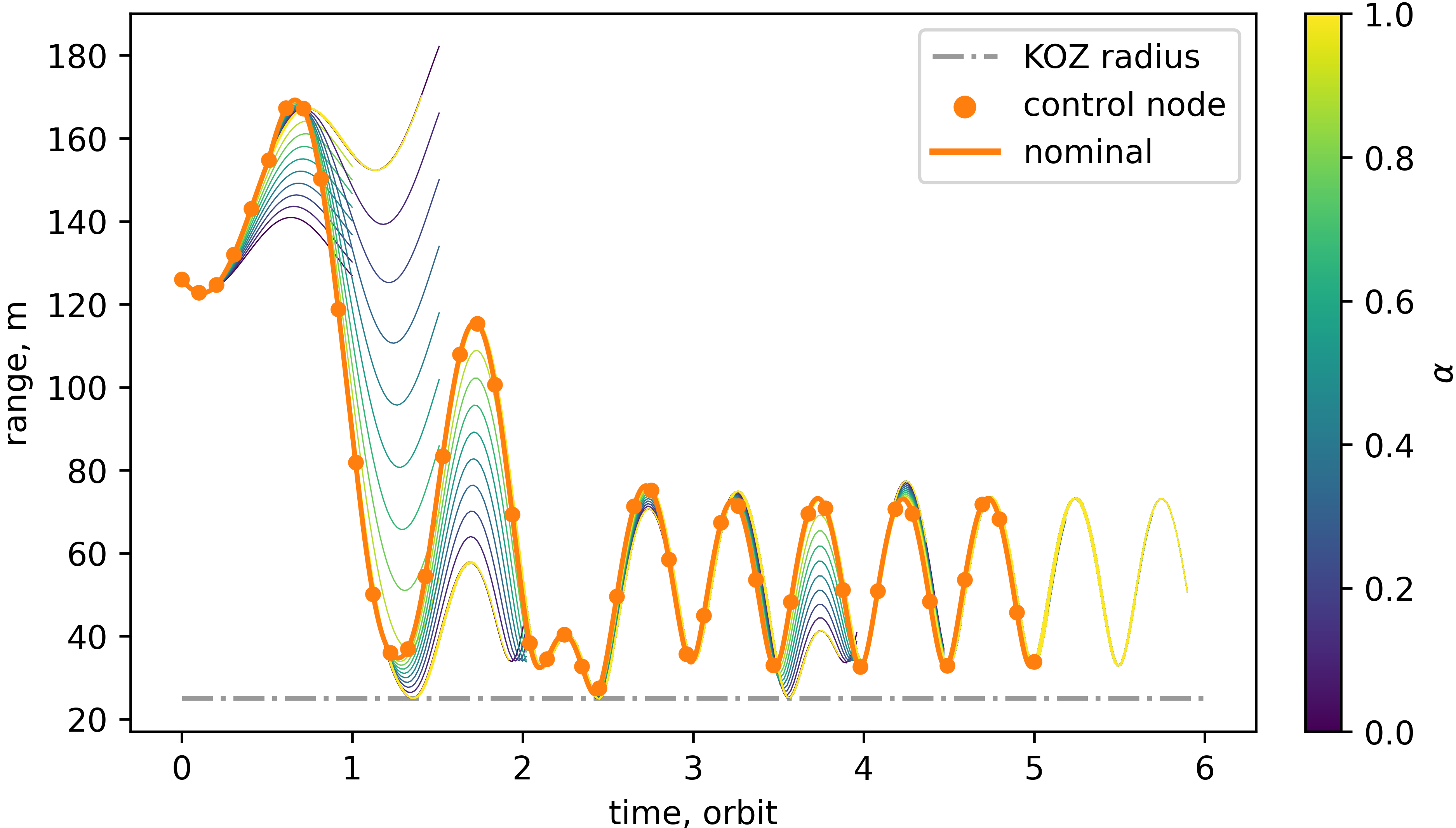}
         \caption{Range from the target}
         \label{fig:ps_ct_range}
     \end{subfigure}
    \caption{\edit{Example trajectory with continuous-time passive safety relative to a 25m spherical keep-out zone (KOZ) under partially imperfect burns.}}
    \label{fig:ps_ct}
\end{figure}

This case study examines a fixed-time trajectory optimization problem under a fixed orbital configuration. 
The transfer duration is set to $t_f = 5$ orbits, and the target spacecraft’s initial orbital elements are fixed at
$\boldsymbol{\alpha}(t_i) = [6738.14~\text{km},\; 5.58\times 10^{-4},\; 51.64^\circ,\; 301.04^\circ,\; 26.18^\circ,\; 68.23^\circ],$
corresponding to a near-circular low Earth orbit. 
The drift time for passive safety is set to $\tau^s = 1$ orbit.
The KOZ is modeled as a sphere of 25\,m radius, and the trajectory is uniformly discretized into $N=50$ nodes.
The maximum control magnitude constraint is omitted because the scenarios considered in the following subsection require only modest maneuvering effort (typically $\Delta V \leq 100\,\mathrm{mm/s}$). 
With ten control nodes per orbit, the impulse per timestep remains well within standard thruster capability, as the required cumulative $\Delta V$ can be realized through a sequence of sufficiently small burns.

The trajectory optimization is solved using SCVx* \cite{oguri2023successive}, an augmented Lagrangian-based SCP algorithm with auto-tuning of the trust region and penalty weights. 
The hyperparameters are provided in Appendix D. 
During the SCP routine, all variables are normalized with respect to the initial guess, thereby scaling each element to the range $[-1,1]$ to enhance numerical stability \cite{malyuta_scp_2022}.
For training dataset generation, the convex relaxation of the problem is first solved by removing the passive-safety constraint in Eq.~\eqref{eq:ocp_rpod_ps_ct} from Eq.~\eqref{eq:ocp_rpod_nonconvex}.
The resulting convex waypoint-hopping trajectory is then used as a warm-start for the SCP.

\begin{figure}[t!]
     \centering
     \begin{subfigure}[h]{\linewidth}
         \centering
         \includegraphics[width=\linewidth]{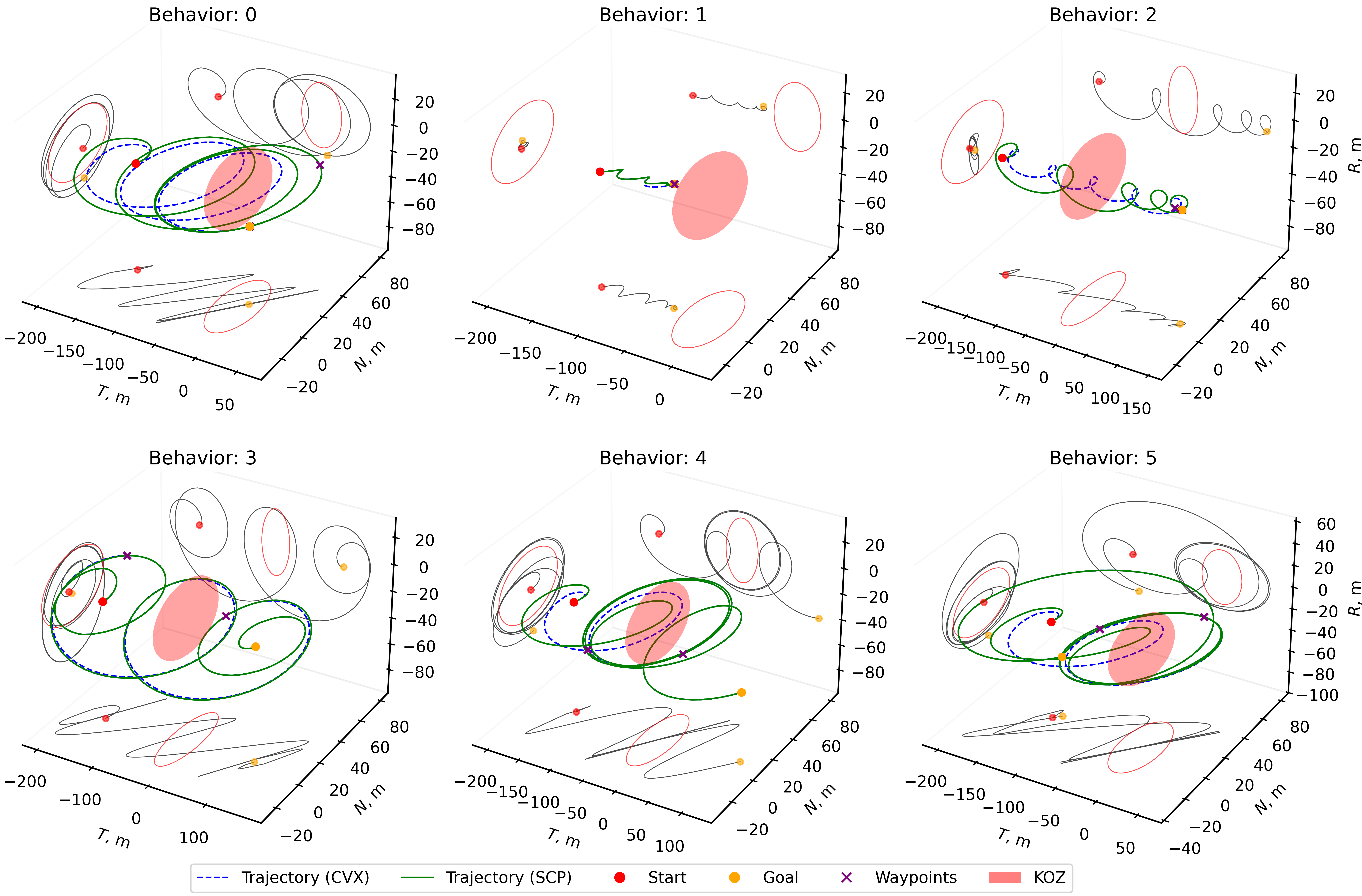 }
         \caption{RTN frame}
         \label{fig:rpod_behav_rtn}
    \end{subfigure} 
    \\
    \begin{subfigure}[h]{\linewidth}
         \centering
         \includegraphics[width=\linewidth]{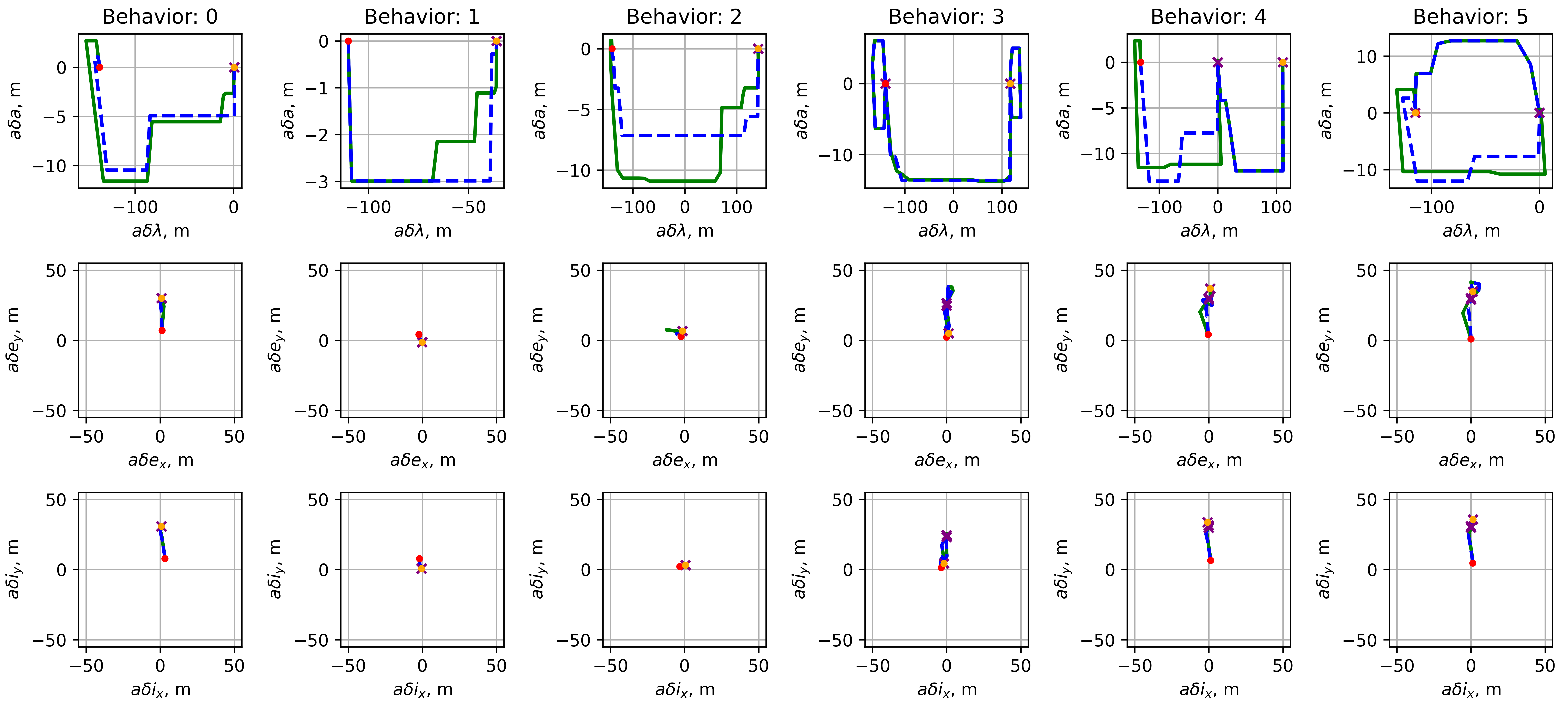}
         \caption{ROE space}
         \label{fig:rpod_behav_roe}
     \end{subfigure}
    \caption{\edit{Representative trajectories corresponding to the behavior modes for the spacecraft proximity operations scenarios in Tables~\ref{tab:wyp_rpod} and ~\ref{tab:commands_rpod}.}}
    \label{fig:rpod_behav}
\end{figure}

With the boundary conditions defined in Table~\ref{tab:wyp_rpod}, representative trajectories provided by solving the above nonconvex trajectory optimization for each behavior mode are illustrated in both the RTN frame and the qnsROE space in Fig.~\ref{fig:rpod_behav}.
For comparison, the solution of convex relaxation that omits the continuous-time passive safety constraint in Eq.~\eqref{eq:ocp_rpod_ps_ct} (denoted as convex waypoint-hopping, or CVX) is also shown in each subfigure. 

\subsubsection{Trajectory Optimization for Refinement after the Test-time Inference}

Similar to the free-flyer scenario, the following optimization problem is solved for the trajectory refinement after the warm-start trajectory is generated:
\begin{subequations} \label{eq:ocp_rpod_feasibility}
\begin{align}
\min_{\{\boldsymbol{x}_k\}_{k=1}^{N}, \{\boldsymbol{u}_k\}_{k=1}^{N}} \quad \eqref{eq:obj_feasibility} \quad 
\text{subject to} \quad 
\eqref{eq:ocp_rpod_dyn}, \eqref{eq:ocp_rpod_ps_ct}, \eqref{eq:ocp_rpod_bc}, 
\end{align}
\end{subequations}
where $\lambda = 0.1$ is defined as a weighting hyperparameter in the objective in Eq.~\eqref{eq:obj_feasibility}.
Note again that the terminal waypoint is determined based on only an input command, so the terminal state constraint is defined using the one from the warm-start trajectory generated from the transformer. 
Furthermore, the waypoint passage constraint in Eq.~\eqref{eq:ocp_rpod_wyp} is not enforced in the problem, as it is assumed that structured waypoint knowledge is not available during the deployment. 

\section{Results} \label{sec:results}

This section evaluates the performance of the proposed SAGES framework through a series of numerical experiments.
The proposed framework is designed to enable text-based commanding for future spacecraft autonomy.
To assess its effectiveness, two Research Questions (RQs) are defined:
\edit{(RQ1) Do the trajectories generated by the transformer exhibit favorable performance in terms of safety and semantic correctness?}, 
\edit{(RQ2) Given the high-quality warm-starts from the transformer, does SAGES generate safe and semantically correct trajectories with a high success rate?}. 
The remainder of this section addresses these research questions to examine the performance and generalization capability of the proposed language-driven safe trajectory generation pipeline.

For the frozen text encoder, {distilbert-base-uncased} model \edit{(67 million parameters)} \cite{sanh2019distilbert} is adopted \edit{based on the consideration of using a lightweight LLM under the strict hardware constraint.}
The hyperparameters of the transformer are summarized in Appendix E.
A total of approximately $N_{\text{data}} \approx 80{,}000$ and $90{,}000$ command–trajectory pairs are used to train the models for free-flyer and spacecraft proximity operation scenarios, respectively (see \edit{Appendix A} for details of dataset generation).

For evaluation, a separate test set of 1,000 unseen command–scenario pairs is randomly sampled. 
In this test set, the boundary conditions are drawn from the same underlying distributions as in training but are not encountered during training.
To assess robustness to variations in natural-language inputs, two command sets are considered:
\begin{enumerate}
\item \edit{Seen commands (templates) with unseen scenarios:} The initial conditions and mission goals are in-distribution but unseen during training. The commands (templates) are identical to those used in training; however, each command instance contains new numerical values, making the full command-scenario pair unseen by the model.
\item \edit{Unseen commands (templates) with unseen scenarios:} Both the initial conditions and the natural-language commands (templates) are not encountered during training.
\end{enumerate}

All computations in this section are performed on an NVIDIA Jetson AGX Orin 64 GB, an embedded GPU platform that integrates an Arm Cortex-A78AE v8.2 64-bit CPU and an NVIDIA Orin GPU.

\subsection{Performance of Warm-Start Trajectories}

\edit{
It is of interest to first investigate the transformer's trajectory generation capability without the SCP refinement. 
To this end, the statistical performance of the Warm-Start (WS) trajectories of SAGES, denoted as SAGES-WS, is first investigated. }
This subsection addresses two key questions related to (RQ1):
\edit{(i) To what extent do the generated warm-start trajectories satisfy the prescribed semantic specifications?}
\edit{(ii) Relative to the convex baseline, do the generated trajectories exhibit improved safety, reflecting a better understanding of the nonconvex constraint landscape?}
To answer these questions, each generated trajectory is evaluated along semantic correctness and violations of nonconvex safety constraints.

For comparison, the convex-based benchmark is also provided. 
In the free-flyer scenario, the CVX stage solves the SCP problem without enforcing the nonconvex collision avoidance constraint in Eq.~\eqref{eq:ocp_ff_obs}; instead, semantic correctness is induced through the designed waypoint, which shapes the resulting trajectory.
In contrast, the spacecraft proximity operation scenario solves the convex waypoint-hopping constraint as CVX; trajectories generated by CVX are always semantically correct through the definition of the designed waypoint.
\edit{Regardless, it is important to note that the two approaches operate under different levels of problem specification. SAGES-WS generates trajectories from only the initial state and a high-level command, implicitly inferring terminal conditions and intermediate structure. 
In contrast, the CVX-based approach requires explicit terminal states and, often, waypoints, thereby solving a fully specified and more constrained problem. As a result, the comparison is not strictly equivalent: the CVX formulation benefits from additional prior information that reduces the feasible search space. Consequently, the evaluation is favorable to the CVX.}
For CVX, the terminal state is taken from the transformer-generated trajectory (SAGES-WS) under the input command with unseen templates. 

\begin{figure}[t]
     \centering
     \begin{subfigure}[th!]{0.85\linewidth}
         \centering
         \includegraphics[width=\linewidth]{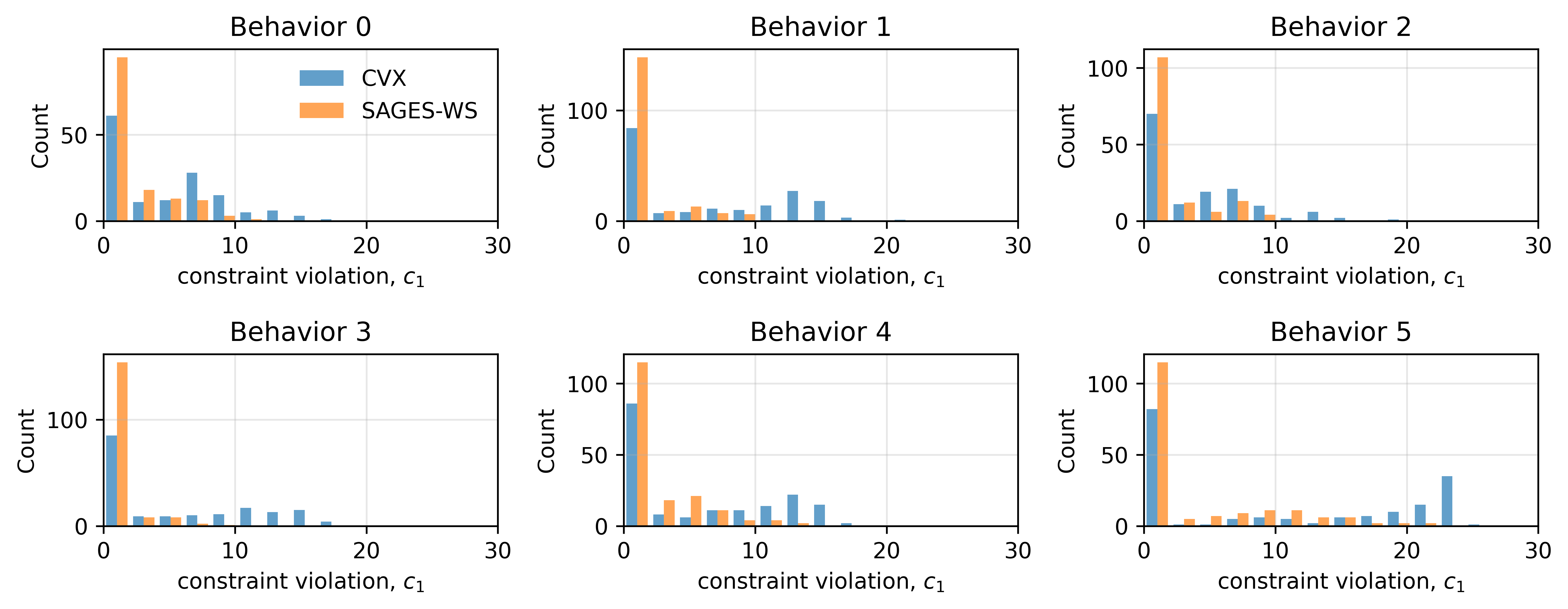 }
         \caption{Free-flyer}
         \label{fig:constr_ws_traj_ff}
    \end{subfigure} 
    \\
    \begin{subfigure}[th!]{0.85\linewidth}
         \centering
         \includegraphics[width=\linewidth]{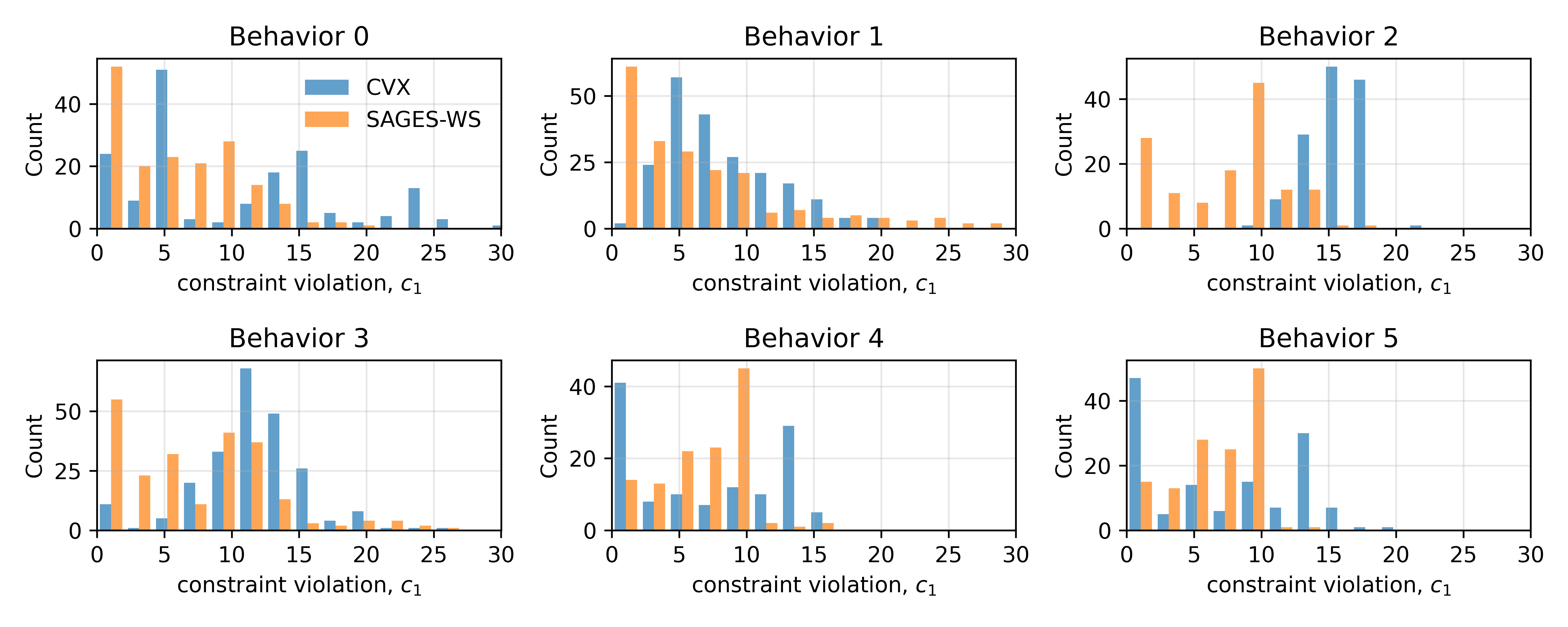}
         \caption{Spacecraft proximity operations.}
         \label{fig:constr_ws_traj_rpod}
     \end{subfigure}
    \caption{\edit{Constraint violations of the CVX and SAGES-WS trajectories across different behavior modes.}}
    \label{fig:constr_ws_traj}
\end{figure}

Using the definitions of semantic correctness in Sec.~\ref{sec:ff_behav_pattern} and Sec.~\ref{sec:rpod_behav_pattern}, Tables~\ref{tab:summary_ff_pre_scp} and~\ref{tab:summary_rpod_pre_scp} summarize statistical semantic correctness and safety (feasibility) of the generated trajectories for the free-flyer and spacecraft proximity operation scenarios, respectively. 
$q$ denotes the tolerance factors for the waypoint domain, defined in Eqs.~\eqref{eq:rpod_wyp_dom} and \eqref{eq:rpod_wyp_dom_dlam}.
In these tables, the reported success rate corresponds to the semantic correctness associated with each input command, as defined in the previous section.
Furthermore, Fig.~\ref{fig:constr_ws_traj} presents the distribution of the cumulative constraint violation (cf. Eq.~\eqref{eq:ctg}) of the generated trajectories based on the unseen commands and command templates, denoted as $c_1$. 
\begin{table}[t] \small
    \caption{\edit{Statistical safety and semantic correctness in warm-start trajectories (free-flyer).}}
    \centering
    \renewcommand*{\arraystretch}{1}
    \begin{tabular}{ccccccccc}
    \toprule
    \multirow{2}{*}{Method} 
        & \multicolumn{2}{c}{Metric} 
        & \multicolumn{6}{c}{Behavior} \\
    \cmidrule(lr){2-3} \cmidrule(lr){4-9}
     & Command & Type 
       & 0 & 1 & 2 & 3 & 4 & 5  \\
    \midrule
    \multirow{2}{*}{CVX}
        && Safety [\%] &  46.15  &  46.99  & 37.67 & 48.55 & 36.57 &  40.34  \\
        && Semantic correctness [\%] & 100 & 100 & 100 & 100 & 100 & 100  \\
    \midrule
    \multirow{4}{*}{SAGES-WS}
        &\multirow{2}{*}{Seen command} & Safety [\%] & 69.93 & 83.61 & 66.44 & 82.08 & 52.0 & 59.66 \\
        &  & \multirow{1}{*}{Semantic Correctness [\%]} 
            & 93.7 & 100 &  95.2 & 100 & 99.4 & 100 \\
        \cmidrule(lr){2-9}
        &\multirow{2}{*}{Unseen command} & Safety [\%] & 59.15 & 78.14 & 71.83 & 84.97 & 60.0 & 64.20 \\
        &  & \multirow{1}{*}{Semantic Correctness [\%]} 
            & 94.4 & 100 &  95.8 & 100 & 99.4 & 100 \\
    \bottomrule
    \end{tabular}
    \label{tab:summary_ff_pre_scp}
\end{table}
\begin{table}[t] \small
    \caption{\edit{\edit{Statistical safety and semantic correctness in warm-start trajectories (spacecraft proximity operation)}}}
    \centering
    \renewcommand*{\arraystretch}{1}
    \begin{tabular}{cccccccccc}
    \toprule
    \multirow{2}{*}{Method} 
        & \multicolumn{3}{c}{Metric} 
        & \multicolumn{6}{c}{Behavior} \\
    \cmidrule(lr){2-4} \cmidrule(lr){5-10}
     & Command & Type & $q$ 
       & 0 & 1 & 2 & 3 & 4 & 5  \\
    \midrule
    \multirow{2}{*}{CVX}
        && Safety [\%] &&  0.0  &  0.5  & 0.6 & 2.3 & 0.0 &  0.0  \\
        && Semantic correctness [\%] & 1 & 100 & 100 & 100 & 100 & 100 & 100  \\
    \midrule
    \multirow{8}{*}{SAGES-WS}
        &\multirow{4}{*}{Seen template} & Safety [\%] & & 25.9 & 23.2 & 13.0 & 21.9 & 7.1 & 6.9 \\
        &  & \multirow{3}{*}{Semantic Correctness [\%]} & 1 
            & 89.51 & 63.51 & 99.24 & 17.21 & 77.17 & 72.25 \\
        &&  & 2 
            & 100 & 92.42 & 99.24 & 42.33 & 91.34 & 88.44 \\
        &&  & 3 
            & 100 & 100 & 99.24 & 66.51 & 96.85 & 92.49 \\
        \cmidrule(lr){2-10}
        &\multirow{4}{*}{Unseen template}  & Safety [\%] &&  20.5 & 16.8 &  17.5 & 22.5 & 7.5 & 8.9 \\
        & & \multirow{3}{*}{Semantic Correctness [\%]} & 1 
            & 89.74 & 63.27 & 99.35 & 12.44 & 50.93 & 73.39 \\
        &&& 2 
            & 100 & 94.39 & 99.35 & 36.36 & 88.20 & 92.74 \\
        &&& 3 
            & 100 & 100 & 99.35 & 62.68 & 91.93 & 95.97 \\
    \bottomrule
    \end{tabular}
    \label{tab:summary_rpod_pre_scp}
\end{table}

The free-flyer scenario examines trajectory generation both with seen and unseen commands in the dataset, along with unseen problem scenarios (i.e., initial state and waypoint position). 
The results show that trajectories generated by the transformer exhibit strong semantic correctness across all behavior modes. 
For several difficult initializations, particularly when the free-flyer’s starting position lies along the Line-of-Sight (LOS) to the goal or to the added waypoint, the CVX warm-start often becomes infeasible, with the issue appearing most prominently in Behaviors 4 and 5 (cf. Fig.~\ref{fig:constr_ws_traj_ff}). 
In contrast, the trajectories of SAGES-WS have a significantly higher likelihood of producing safe and semantically correct trajectories without requiring additional refinement.
Also, according to Fig.~\ref{fig:constr_ws_traj_ff}, the transformer produces consistently lower violation magnitudes across all behaviors, with a large mass concentrated at zero, indicating that the majority of warm-start trajectories are already achieving collision avoidance.

In the spacecraft proximity operations scenario, under challenging continuous-time passive safety conditions with imperfect burns, the convex waypoint-hopping solutions (CVX) rarely satisfy the safety constraint, even when the waypoints themselves are passively safe. 
In contrast, SAGES-WS trajectories generated by the transformer improve the likelihood of producing solutions that are already passively safe without additional refinement up to 25\% (Behavior 0), while maintaining quantitative semantic correctness when a moderate waypoint tolerance (e.g., $q=2$ or $3$) is allowed.
Table~\ref{tab:summary_rpod_pre_scp} further shows that performance is comparable under both seen and unseen command templates. 
This indicates that the model does not overfit to particular textual patterns; instead, it encodes the underlying semantic content, including numerical quantities, within the latent space of the text encoder.
Figure~\ref{fig:constr_ws_traj_rpod} highlights this trend in the distribution of constraint violations. For Behaviors~0--2, SAGES-WS lowers the violation distribution while preserving semantic correctness, with most trajectories remaining within the waypoint set for $q \leq 2$. 
A trade-off arises for Behaviors~3--5, where the violation distribution is reduced but at the cost of lower semantic-correctness rates. 
In particular, Behavior 3 yields lower semantic correctness for the SAGES-WS trajectory compared to the other behavior modes. 
This occurs because the qnsROE changes rapidly near this waypoint in the relative semi-major axis $\delta a$, as illustrated in Fig.~\ref{fig:rpod_behav_roe}. Consequently, the waypoint set becomes too small at the desired passage epoch, increasing the likelihood that the resulting trajectory is classified as semantically incorrect.
Conversely, for Behaviors~4 and~5, high-level commands are consistently satisfied, although the reduction in violation magnitude is less pronounced; nonetheless, the number of trajectories with large violations (e.g., $c_1 > 10$) is significantly reduced.
This trade-off is consistent with observations in prior work, where tension often arises between optimizing a reward signal (here reflected in semantic correctness) and enforcing hard constraints~\cite{art_ieeeaero24}, and it continues to appear even in multi-objective formulations~\cite{takubo2026agile}. 
Overall, the results show that the proposed framework captures the intended semantics of the command and generates trajectories that exhibit more favorable safety and correctness properties than benchmark waypoint-hopping trajectories. 

\subsection{Performance of Converged Solutions in Sequential Convex Programming}

In this subsection, the performance of SAGES is evaluated based on the refined trajectory after the SCP.
To address (RQ2), two key questions of interest are: \edit{(i) Does the usage of the trajectories generated by SAGES provide an advantageous property in algorithmic performance of the SCP?}, and \edit{(ii) Are the converged solutions in the SCP not only safe but also compliant with the high-level command?}
The generated warm-starts are used as a warm-start of the SCP that solves Eqs.~\eqref{eq:ocp_ff_feasibility} and ~\eqref{eq:ocp_rpod_feasibility}, respectively. 
The converged solutions are then compared against the benchmark SCP solutions warm-started by the convex waypoint-hopping (CVX) solutions. 

\begin{figure}[t]
     \centering
     \begin{subfigure}[th!]{0.9\linewidth}
         \centering
         \includegraphics[width=\linewidth]{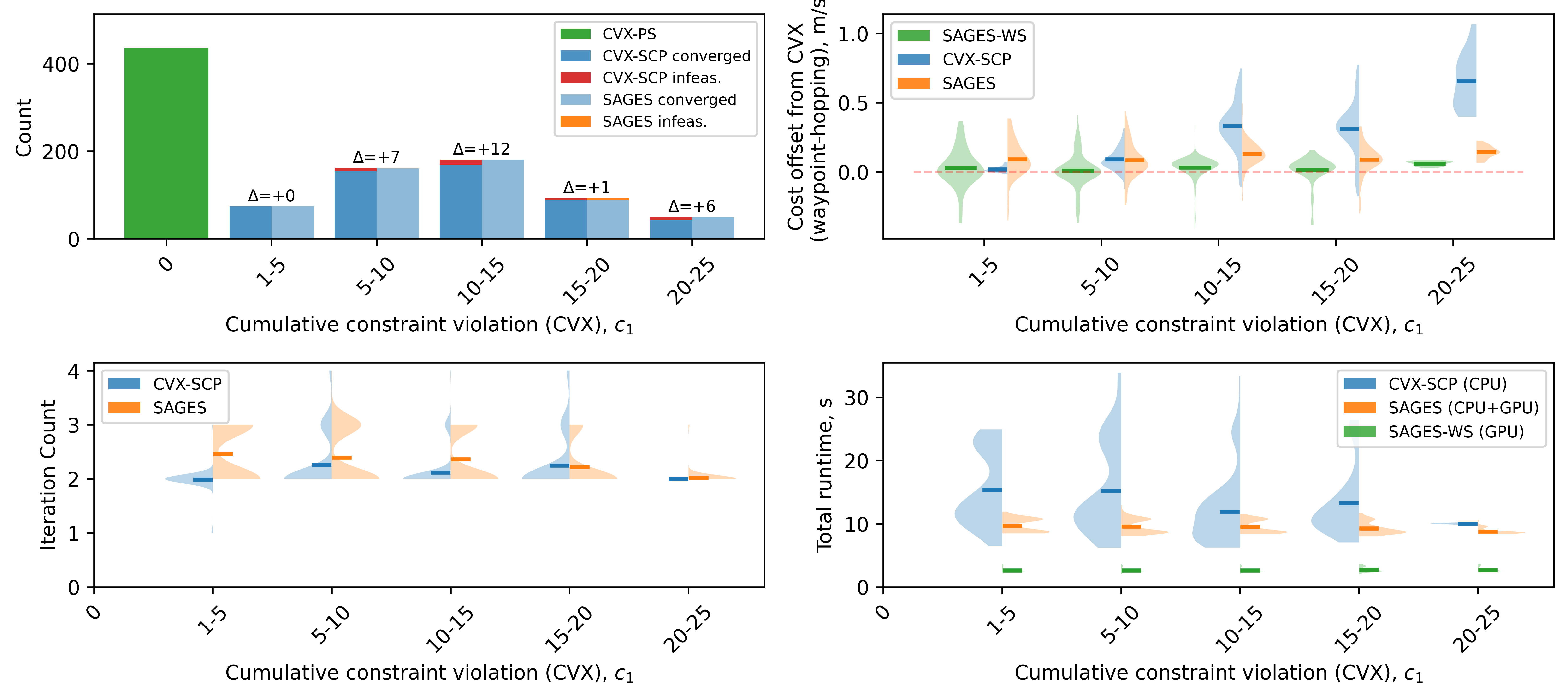 }
         \caption{Free-flyer}
         \label{fig:ws_analysis_FF}
    \end{subfigure} \\
    \begin{subfigure}[th!]{0.9\linewidth}
         \centering
         \includegraphics[width=\linewidth]{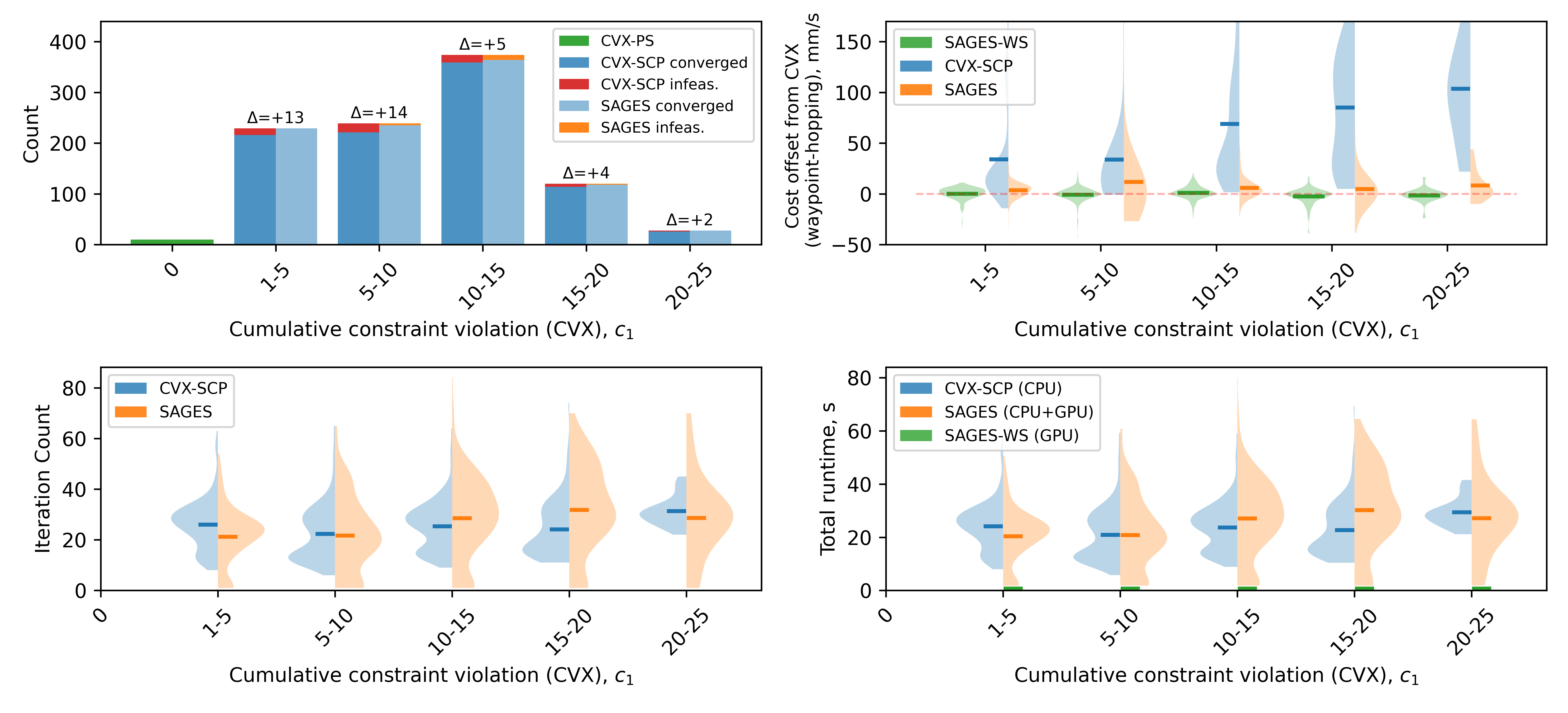}
         \caption{Spacecraft proximity operations}
         \label{fig:ws_analysis_rpod}
     \end{subfigure}
    \caption{\edit{Comparison of SCP performance using warm-starts from SAGES-WS and a CVX baseline.}}    
    \label{fig:ws_analysis}
\end{figure}

First, Figure~\ref{fig:ws_analysis} summarizes four statistical metrics across statistical probability of convergence in the SCP, converged objective value (as the cost offset from the convex waypoint-hopping solution), iteration counts of the SCP until convergence, and the total run-time, which is the sum of the SCP routine and the warm-starting.
These values for both the free-flyer and spacecraft proximity operations scenario are summarized as a histogram and a series of violin plots.
\edit{This involves 1,000 unseen, in-distribution test scenarios, and the abscissa is binned across the cumulative constraint violation of the CVX solution.}
Some problem scenarios satisfy the nonconvex constraints using a purely convex waypoint-hopping solution (i.e., $c_1=0$), shown as the green bar in the top-left plot of Fig.~\ref{fig:ws_analysis}). 
These cases do not require SCP refinement and are therefore excluded from the statistical analysis. 
The reported distributions of objective optimality, SCP iteration count, and total run-time are computed only over scenarios in which SCP converges under the corresponding warm-starting method.
Furthermore, the central question for SAGES is whether the SCP-refined trajectories remain semantically correct with the high-level textual commands. 
To evaluate this, Tables~\ref{tab:summary_ff_post_scp} and \ref{tab:summary_rpod_post_scp} report both the SCP convergence rate (for warm-starting using CVX and SAGES-WS) and the semantic correctness of the resulting solutions.
\edit{Note that $q$ denotes the tolerance factors for the waypoint domain in Eqs.~\eqref{eq:rpod_wyp_dom} and \eqref{eq:rpod_wyp_dom_dlam}.}
For comparison, the tables also include a fuel-optimal solution (cf. Eq.~\eqref{eq:ocp_rpod_obj}) initialized with the SAGES-WS.

\begin{table}[t] \small
    \caption{\edit{Statistical SCP convergence rate and semantic correctness of the converged trajectories (free-flyer).}}
    \centering
    \renewcommand*{\arraystretch}{1}
    \begin{tabular}{ccccccccc}
    \toprule
    \multirow{2}{*}{Method} 
        & \multicolumn{2}{c}{Metric} 
        & \multicolumn{6}{c}{Behavior} \\
    \cmidrule(lr){2-3} \cmidrule(lr){4-9}
     & Command & Type 
       & 0 & 1 & 2 & 3 & 4 & 5  \\
    \midrule
    \multirow{2}{*}{CVX-SCP}
        & \multirow{2}{*}{\makecell[c]{Feasibility \\ ($\lambda = 1$)}} 
            & SCP Convergence (Safety) [\%] &  93.84  &  96.72  & 94.56 & 100 & 96.0 &  95.45  \\
         & & Semantic correctness [\%] & 95.8 & 96.7 & 97.2 & 99.4 & 96.0 & 100  \\
    \midrule
    \multirow{4}{*}{SAGES}
        &\multirow{2}{*}{\makecell[c]{Feasibility \\ ($\lambda = 1$)}} & SCP Convergence (Safety) [\%] & 99.32 & 100 & 98.64 & 100 & 98.86 & 100\\
        &  & \multirow{1}{*}{Semantic Correctness [\%]} 
            & 95.1 & 100 & 97.9 & 100 & 98.9 & 100 \\
        \cmidrule(lr){2-9}
        &\multirow{2}{*}{Fuel Optimality} & SCP Convergence (Safety) [\%] & 99.32 & 100 & 98.64 & 100 & 98.86 & 100\\
        &  & \multirow{1}{*}{Semantic Correctness [\%]} 
            & 0.0 & 13.7 & 0.0 & 16.8 & 0.0 & 100 \\
    \bottomrule
    \end{tabular}
    \label{tab:summary_ff_post_scp}
\end{table}
\begin{table}[t] \small
    \caption{\edit{Statistical SCP convergence rate and semantic correctness of the converged trajectories (spacecraft proximity operation). }}
    \centering
    \renewcommand*{\arraystretch}{1}
    \begin{tabular}{cccccccccc}
    \toprule
    \multirow{2}{*}{Method} 
        & \multicolumn{3}{c}{Metric} 
        & \multicolumn{6}{c}{Behavior} \\
    \cmidrule(lr){2-4} \cmidrule(lr){5-10}
     & Command & Type & $q$ 
       & 0 & 1 & 2 & 3 & 4 & 5  \\
    \midrule
    \multirow{4}{*}{CVX-SCP}
    & \multirow{4}{*}{\makecell[c]{Feasibility \\ ($\lambda = 0.1$)}} 
    & SCP Convergence (Safety) [\%] &
        & 90.38 & 91.84 & 94.16 & 85.17 & 95.03 & 85.48 \\
    && \multirow{3}{*}{Semantic Correctness [\%]} & 1 
        & 30.77 & 28.57 & 27.92 & 0.00 & 38.51 & 58.06 \\
    &&& 2
        & 62.18 & 60.20 & 58.44 & 5.74 & 66.46 & 70.16 \\
    &&& 3
        & 71.15 & 71.94 & 67.53 & 20.10 & 75.16 & 75.00 \\
    \midrule
    \multirow{8}{*}{SAGES}
        & \multirow{4}{*}{\makecell[c]{Feasibility \\ ($\lambda = 0.1$)}}  
        & SCP convergence [\%] &
            & 100 & 93.88 & 96.10 & 100 & 100 & 100 \\
        &&\multirow{3}{*}{Semantic Correctness [\%]}& 1
            & 89.74 & 62.76 & 95.45 & 12.44 & 50.93 & 73.39 \\
        &&& 2
            & 100 & 89.80 & 95.45 & 36.36 & 87.58 & 92.74 \\
        &&& 3
            & 100 & 92.35 & 95.45 & 62.20 & 91.93 & 95.97 \\
        \cmidrule(lr){2-10}
        & \multirow{4}{*}{\makecell[c]{Fuel Optimality }} 
        & SCP convergence (Safety) [\%] &
            & 78.21 & 85.20 & 85.06 & 65.07 & 77.64 & 48.39 \\
        &&\multirow{3}{*}{Semantic Correctness [\%]} & 1
            & 26.28 & 35.20 & 51.95 & 0.00 & 0.00 & 0.00 \\
        &&& 2 
            & 39.10 & 65.31 & 68.18 & 0.00 & 0.00 & 0.00 \\
        &&& 3 
            & 49.36 & 78.57 & 80.52 & 0.00 & 0.00 & 0.00 \\        
    \bottomrule
    \end{tabular}
    \label{tab:summary_rpod_post_scp}
\end{table}

For the free-flyer scenario, all results use unseen text commands paired with unseen problem scenarios.
Figure~\ref{fig:ws_analysis_FF} shows that trajectories generated by SAGES consistently exhibit lower fuel consumption than the CVX-SCP baselines.
The advantage is most pronounced in scenarios where the CVX initialization incurs significant constraint violations due to challenging waypoint geometries that make passive safety difficult to satisfy. 
In such cases, the transformer produces semantically correct trajectories that are already close to passive safety, requiring only modest corrections during the subsequent SCP stage.
As a result, the converged objective is typically lower for SAGES, since the CVX warm-start must undergo large corrective updates to achieve feasibility, which often leads to less favorable convergence behavior.
Furthermore, SAGES exhibits a significantly shorter total run-time compared to the CVX-SCP, despite the similar iteration counts in the SCP, which is almost always less than four iterations. 
This overhead in the total run-time arises because the CVX baseline must solve an additional SCP to construct its warm-start, adding roughly 5–7 s per instance. 
In contrast, the inference time by the transformer model is effectively negligible.
Table~\ref{tab:summary_ff_post_scp} further shows that, with the feasibility-based projector, SAGES yields a higher rate of safe and convergent solutions while maintaining strong semantic alignment with the input text.
The fuel-optimality variant also produces safe trajectories but places less emphasis on semantic fidelity, occasionally deviating from the commanded maneuver pattern and getting close to the central traverse behavior pattern near the straight line. 
Overall, SAGES provides safer and more semantically correct trajectories, while simultaneously enabling more fuel-efficient solutions, compared to the standard CVX-SCP warm-start strategy.

For the spacecraft proximity operations scenario, the results are obtained by using unseen command templates paired with unseen problem scenarios. 
Overall, the SCP exhibits trends consistent with the free-flyer results. 
As shown in Fig.~\ref{fig:ws_analysis_rpod}, warm-starting SCP with the SAGES-WS improves convergence, even when the convex waypoint-hopping (CVX) initialization presents non-negligible constraint violations. 
The SAGES solutions also reliably recover the nominal CVX control cost, with most cases falling within 20 mm/s below to 50 mm/s above the CVX objective. 
By contrast, the CVX-SCP solution often incurs significantly higher fuel usage, sometimes exceeding an additional 100 mm/s, particularly when the CVX warm-start is highly infeasible ($c_1>15$). 
Despite these differences, the iteration count and total run-time remain comparable to, or better than, those of CVX-SCP. 
The inference time of the transformer on the embedded GPU is negligible; instead, the run-time is dominated by the high-dimensional SCP, whose cost grows with the number of time discretization nodes. 
Although time-dilation methods~\cite{malyuta_scp_2022} could reduce this burden, they increase sensitivity to initialization, especially for impulsive systems~\cite{kumagai2023robust}.
Since control profiles in the spacecraft proximity operations scenario tend to exhibit impulsive behavior in the training dataset, a practical and computationally efficient heuristic is to: (i) generate a finely discretized trajectory using the transformer, (ii) select the $M$ time nodes corresponding to the largest control magnitudes, and (iii) solve a reduced-time-grid SCP formulation in which control is applied only at the selected epochs.
Table~\ref{tab:summary_rpod_post_scp} further shows that feasibility-based refinement yields high convergence rates ($\simeq 95\%$ or more across all behavior modes) and strong post-refinement semantic correctness ($\geq90\%$ for all modes except Behavior 3, which is already challenging at warm-start; cf. Table~\ref{tab:summary_rpod_pre_scp}). 
A clearer contrast emerges when comparing feasibility-based and fuel-optimal solves initialized with the same SAGES-WS solutions: fuel-optimal SCP suffers markedly lower convergence, as the semantically structured warm-starts are generally far from the minimum-fuel solution and require large corrective steps that often lead to nonconverging behaviors. 
Even when fuel-optimal SCP converges, semantic correctness deteriorates sharply, where Behaviors 3–5 often skip the intended waypoints entirely. 
These results indicate that feasibility-oriented refinement is substantially better aligned with preserving the command-driven behavioral structure, enabling reliable, semantically correct, and safe trajectories.

\subsection{Qualitative Comparison of the Representative Trajectories}

\begin{figure}[th!]
     \centering
     \begin{subfigure}[th!]{0.49\linewidth}
         \centering
         \includegraphics[width=\linewidth]{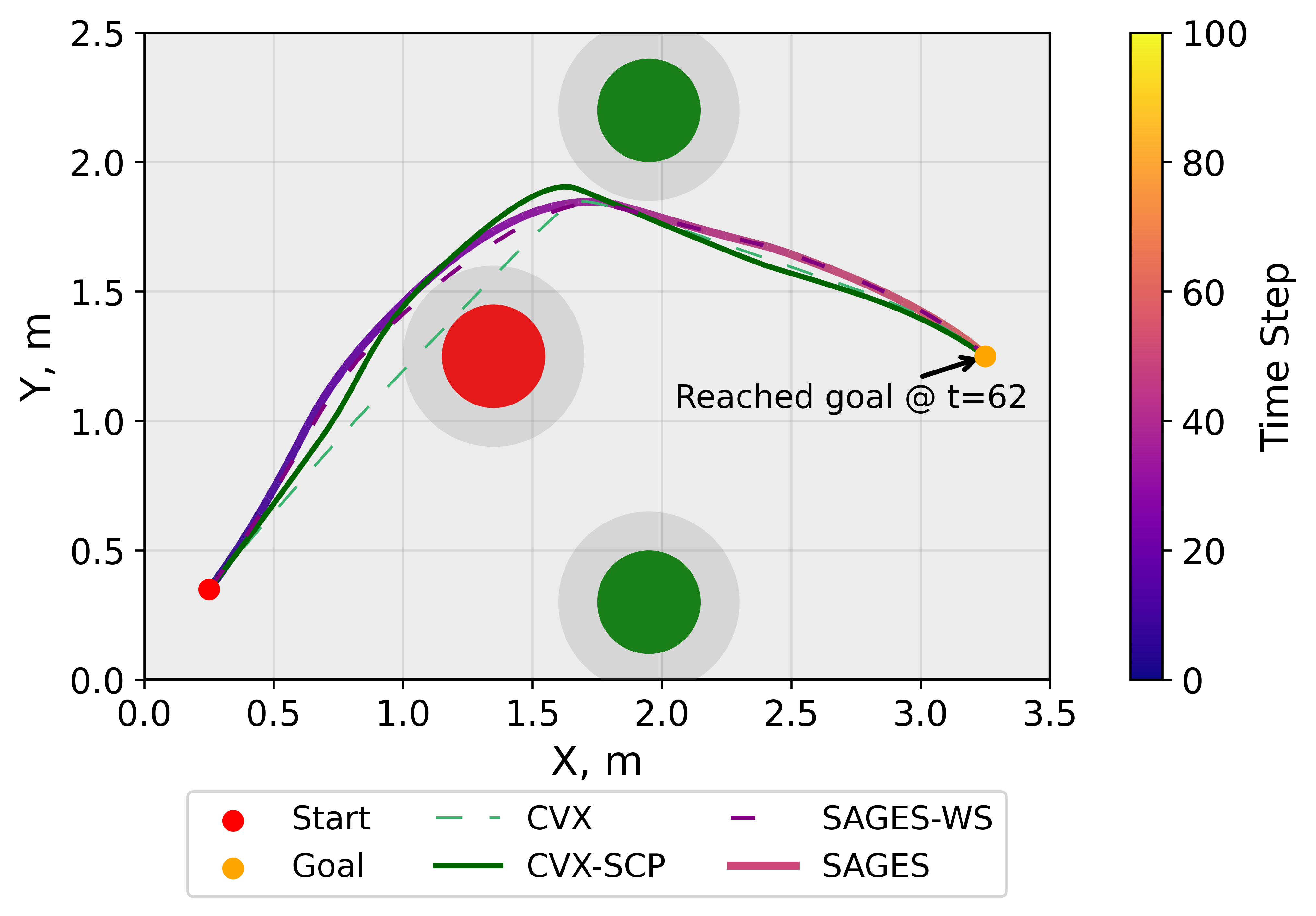 }
         \caption{\edit{Free-flyer Behavior 0 (fast left passage).}}
         \label{fig:rep_traj_ff_1}
    \end{subfigure} 
    \begin{subfigure}[th!]{0.49\linewidth}
         \centering
         \includegraphics[width=\linewidth]{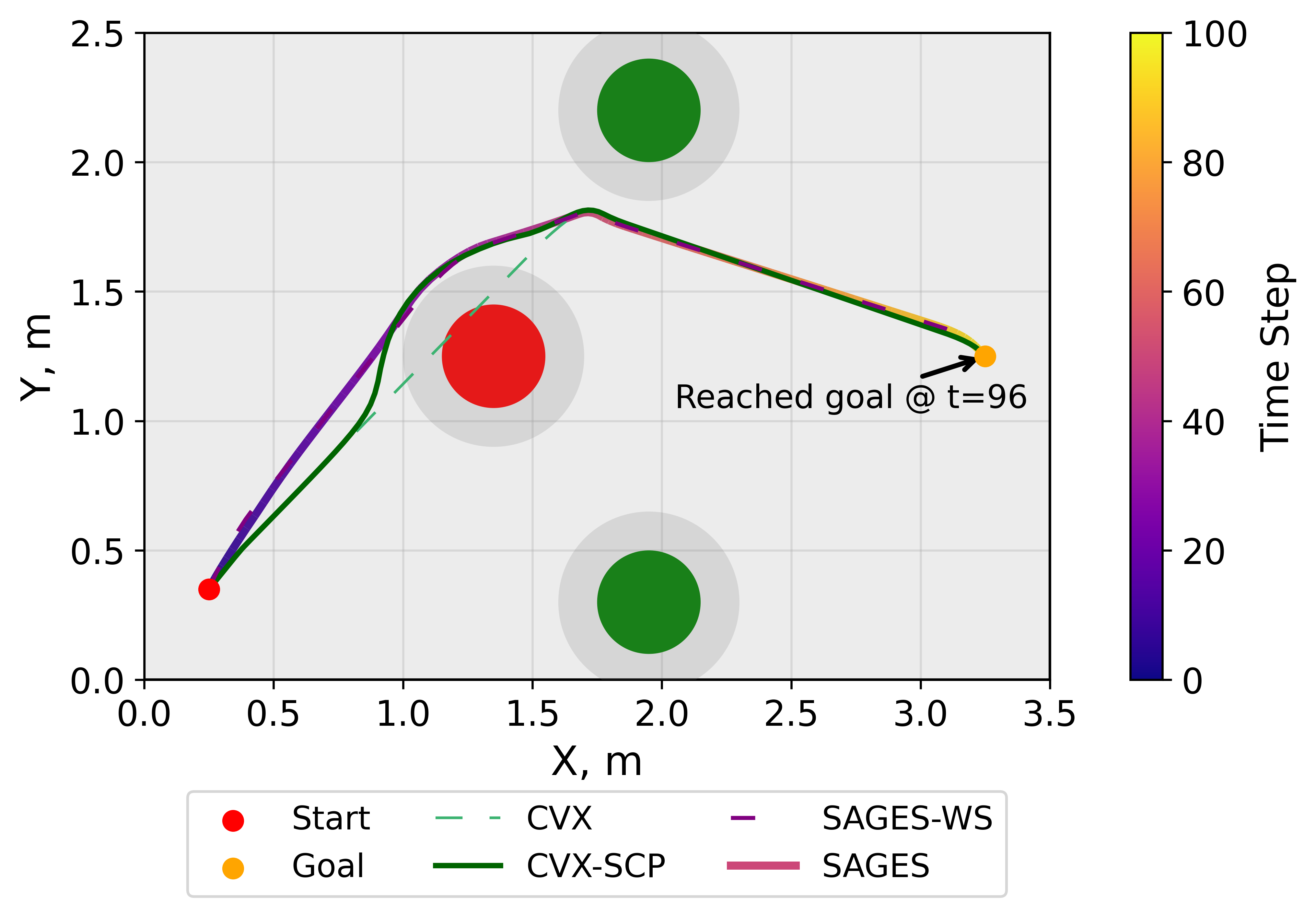}
         \caption{\edit{Free-flyer Behavior 1 (slow left passage).}}
         \label{fig:rep_traj_ff_2}
    \end{subfigure} \\
    \begin{subfigure}[th!]{0.99\linewidth}
         \centering
         \includegraphics[width=\linewidth]{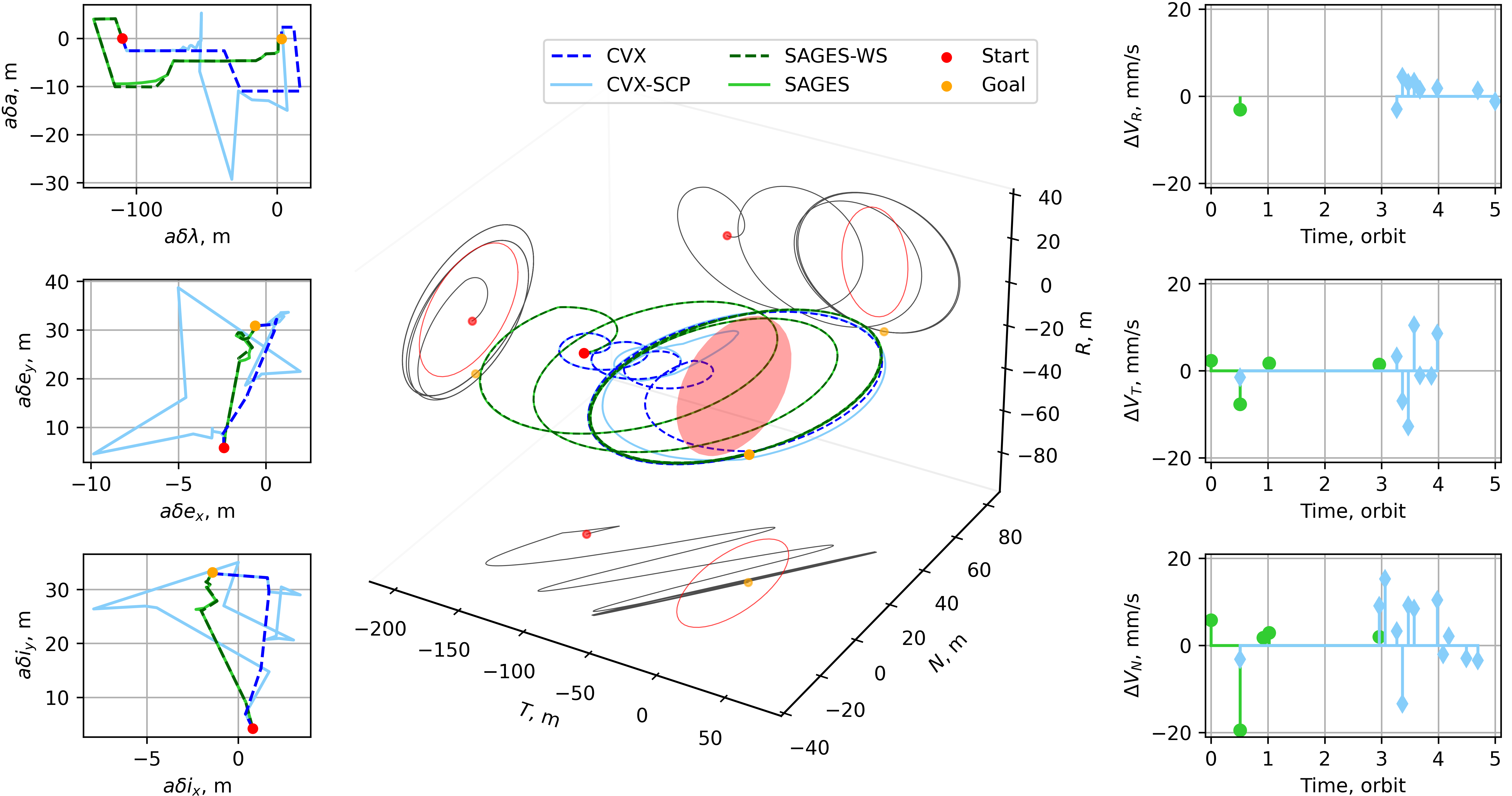}
         \caption{\edit{Spacecraft proximity operations Behavior 0 (approach to the target and circumnavigate). }}
         \label{fig:rep_traj_rpod}
    \end{subfigure}
    \caption{\edit{Representative trajectories generated by SAGES based on transformer-based warm-starting (SAGES-WS) with comparison to CVX-SCP, warm-started with convex-based method (CVX).} }
    \label{fig:rep_traj}
\end{figure}

Figure~\ref{fig:rep_traj} qualitatively compares the representative trajectories generated by SAGES and CVX–SCP.
In the free-flyer scenario, Figs.~\ref{fig:rep_traj_ff_1} and \ref{fig:rep_traj_ff_2} depict two maneuvers initialized from the same state but executing a fast and a slow left passage, respectively, under the commands ``\textit{To preserve clearance, a left-bias routing skirts the KOZ with rapid motion}'' and ``\textit{Perform a broad left-side arc, expanding clearance for extended standoff with minimal \(\Delta v\).}''
The former reaches its terminal waypoint in 24.8 s, while the latter requires 38.4 s; both remain collision-free and accurately reflect their respective high-level behaviors.
By contrast, the CVX initialization produces an initially unsafe trajectory. 
Although the SCP refinement eventually recovers feasibility, the required corrections are substantial, leading to inflated fuel expenditure and a noticeable deviation from the intended motion.

A similar trend is observed in the spacecraft proximity operations scenario.
Starting from $\boldsymbol{x}_{i} = [0,-110,-2.4, 5.8$, $0.8, 4.2]~\mathrm{m}$ under the command \textit{“Approach the target over 4.7 orbits, then circumnavigate while maintaining a safe relative orbit’’} (Behavior~0), the trajectories produced by each initialization method and their SCP-refined counterparts are shown in Fig.~\ref{fig:rep_traj_rpod}, together with the corresponding control profiles.
The CVX warm-start enters the approach phase with inadequate RN-plane separation and only widens the relative orbit immediately before arrival.
This yields a trajectory that is highly passively unsafe, inducing large corrective maneuvers during SCP and consequently increasing the overall control effort.
In contrast, the SAGES-WS generates a safe RN-plane offset from the outset and transitions smoothly into the terminal circumnavigation.
Only a small correction is required to satisfy the continuous-time passive safety constraints, resulting in a final SAGES trajectory that remains nearly indistinguishable from its warm-start.

\subsection{Hardware Demonstration on the Free-flyer Robotic Testbed}\label{subsec:ff_hardware_demo}

To validate the performance of SAGES in the physical world, hardware experiments are conducted using the free-flyer robotic testbed with the NVIDIA Jetson AGX Orin 64 GB platform.
Once the open-loop guidance trajectory is generated by SAGES, a low-level PID controller operating at \(10 ~ \text{Hz}\) with \(K_P = \text{diag}[2.0, 2.0, 0.2]\), \(K_D = \text{diag}[45.0, 45.0, 0.4]\), and \(K_I = \text{diag}[0, 0, 0]\) was employed to track the guidance trajectory. 

\begin{figure}[t]
    \centering
    \begin{subfigure}{0.6\linewidth}
        \centering
        \includegraphics[width=\linewidth]{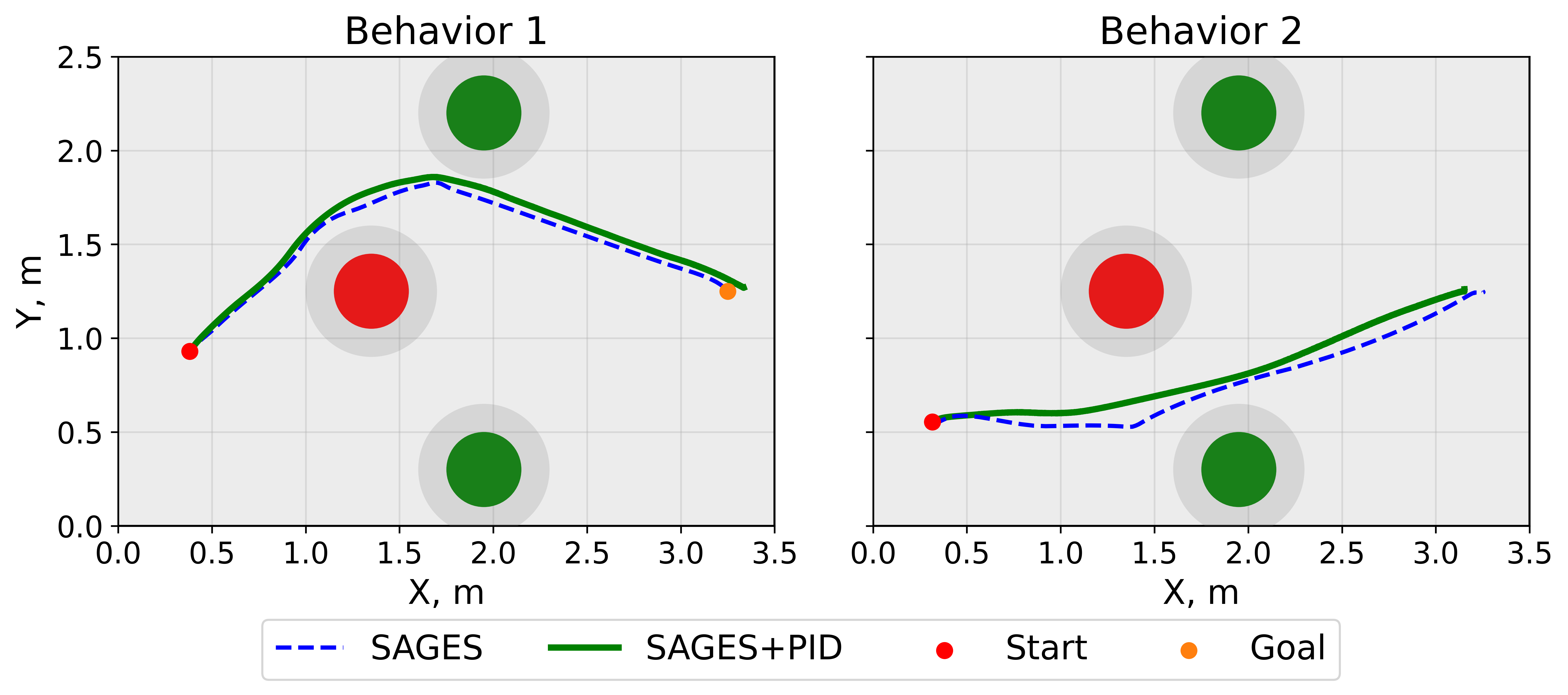}
        \caption{2D-planar trajectory}
        \label{fig:ff_hardware_traj_c}
    \end{subfigure}
    \begin{subfigure}{0.39\linewidth}
        \centering
        \includegraphics[width=\linewidth]{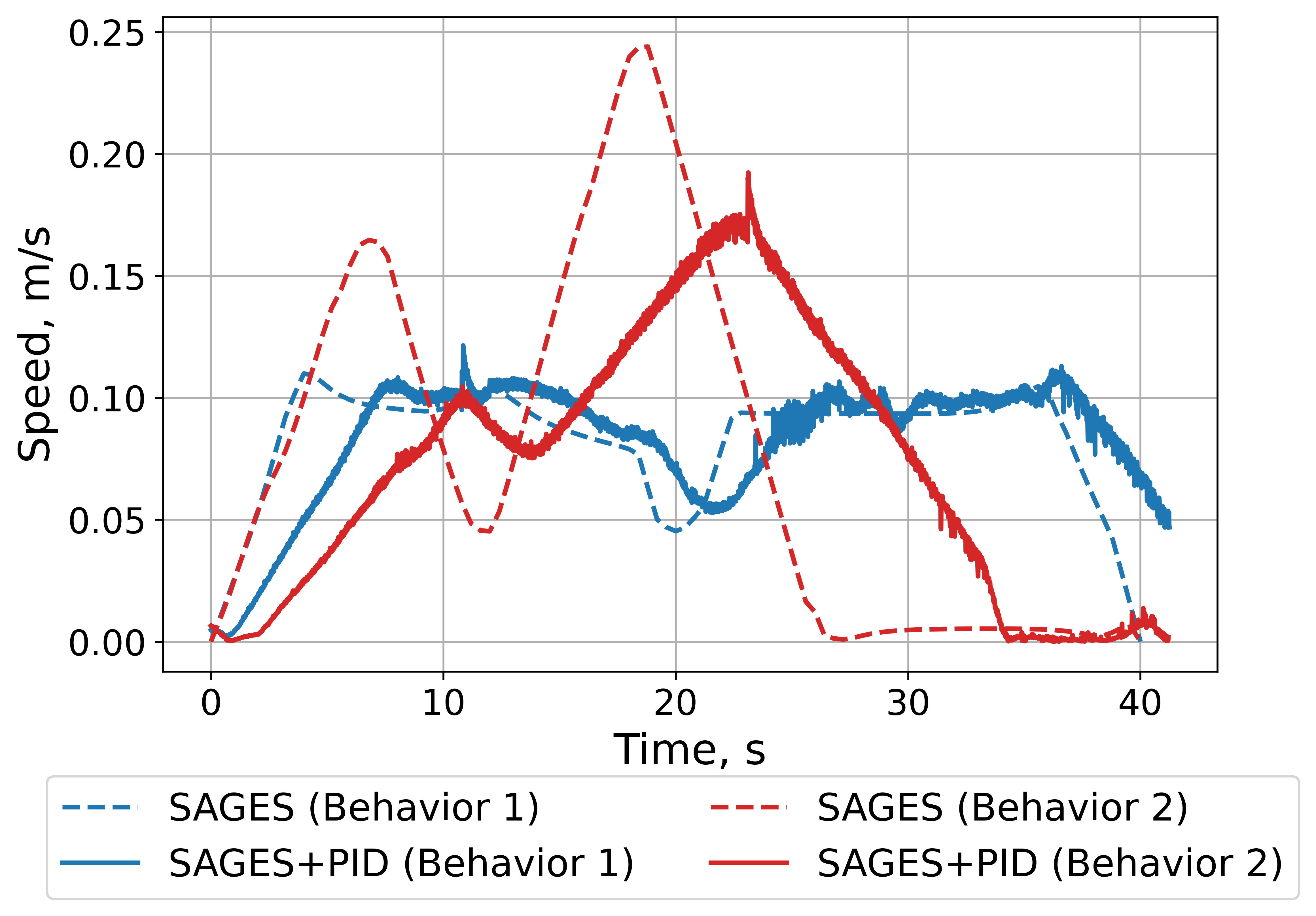}
        \caption{Speed profile}
        \label{fig:ff_hardware_controlP_c}
    \end{subfigure}
    \caption{\edit{Comparison between planned free-flyer trajectories (SAGES) and hardware testbed executions under tracking based on PID control (SAGES+PID).}}
    \label{fig:ff_hardware_traj}
\end{figure}

Figure~\ref{fig:ff_hardware_traj} presents both qualitative and quantitative results from the free-flyer hardware experiments \footnote{The video of the hardware experiment is available at \url{https://semantic-guidance4space.github.io/}}. 
The left panel of Fig.~\ref{fig:ff_hardware_traj_c} illustrates the planned SAGES trajectory and its real-time execution under the PID controller for Behavior 1, corresponding to the command \textit{“For safe passage, skirt the left body while ensuring KOZ compliance and wide standoff.”} 
In contrast, the right panel shows the executed trajectory for the command \textit{“In this profile, a right-bias routing allows rapid transit near the KOZ.”}
On the AGX Orin platform, the CVX warm-start pipeline requires 6–9 s, followed by 5-8 s of SCP refinement.
In comparison, SAGES generates warm-starts in 2-3 s, with the SCP refinement stage taking a similar 5–8 s.
While the refinement time is comparable or slightly better for SAGES relative to the CVX pipeline, the overall advantage is clearly observed in the warm-start phase, where the transformer dramatically outperforms the CVX-based warm-start, and SAGES remains competitive while providing significantly safer and more semantically correct trajectories.
This setup demonstrates the real-world applicability of the proposed approach under realistic computational constraints.

The distinction between fast and slow behavior is illustrated in Fig.~\ref{fig:ff_hardware_controlP_c}, which shows the time history of the vehicle’s speed in the guidance (reference) trajectory and the realized (tracking) trajectory using the hardware. 
The slow behavior (Behavior~1) maintains a nonzero velocity until the final quarter of the horizon (30–40 s), whereas the fast behavior (Behavior~2) decelerates to near zero by approximately 33 s. 
Furthermore, the cumulative thruster firing times are 89.7 s for Behavior~1 and 131.0 s for Behavior~2, which confirms the increased fuel expenditure required to reach the goal more quickly.
Discrepancies between the reference and tracked velocity profiles arise primarily from modeling errors and from the damping characteristics of the PID controller during reference overshoot.
Nevertheless, the hardware experiment demonstrates that the vehicle can reliably realize the intended velocity profile specified through natural-language commands using SAGES.

\subsection{Out-of-Distribution Detection in the Latent Space}

Throughout this paper, it is assumed that there exists a discrete and finite set of behavior modes that express the spacecraft's behavior. 
This assumption can be effectively leveraged for building safeguards and run-time monitors to perform out-of-distribution (OOD) detection around \edit{SAGES} (cf. Fig.~\ref{fig:summary}).

\begin{figure}[t]
    \centering
    \begin{subfigure}{0.45\linewidth}
        \centering
        \includegraphics[width=\linewidth]{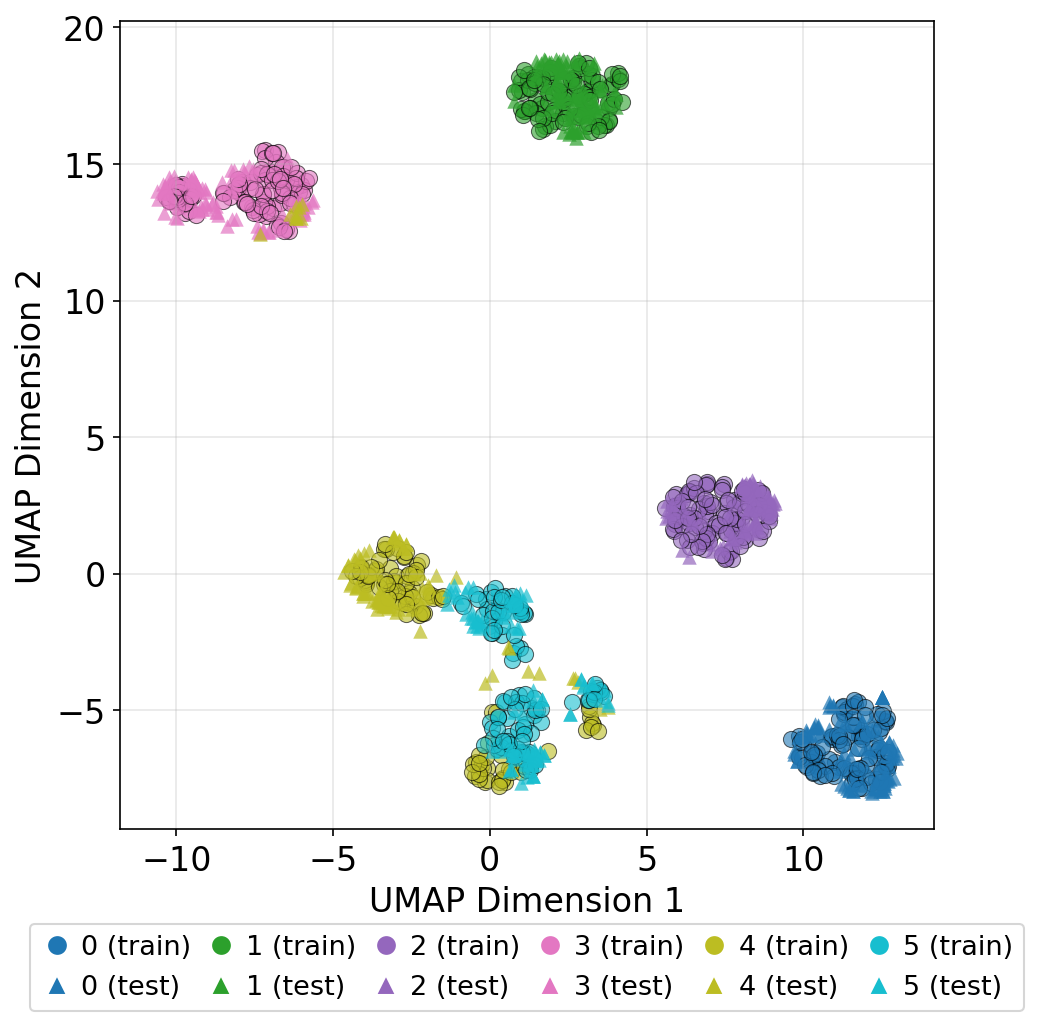}
        \caption{\edit{Two-dimensional UMAP of the input commands.}}
        \label{fig:txt_umap}
    \end{subfigure}
    \begin{subfigure}{0.52\linewidth}
        \centering
        \includegraphics[width=\linewidth]{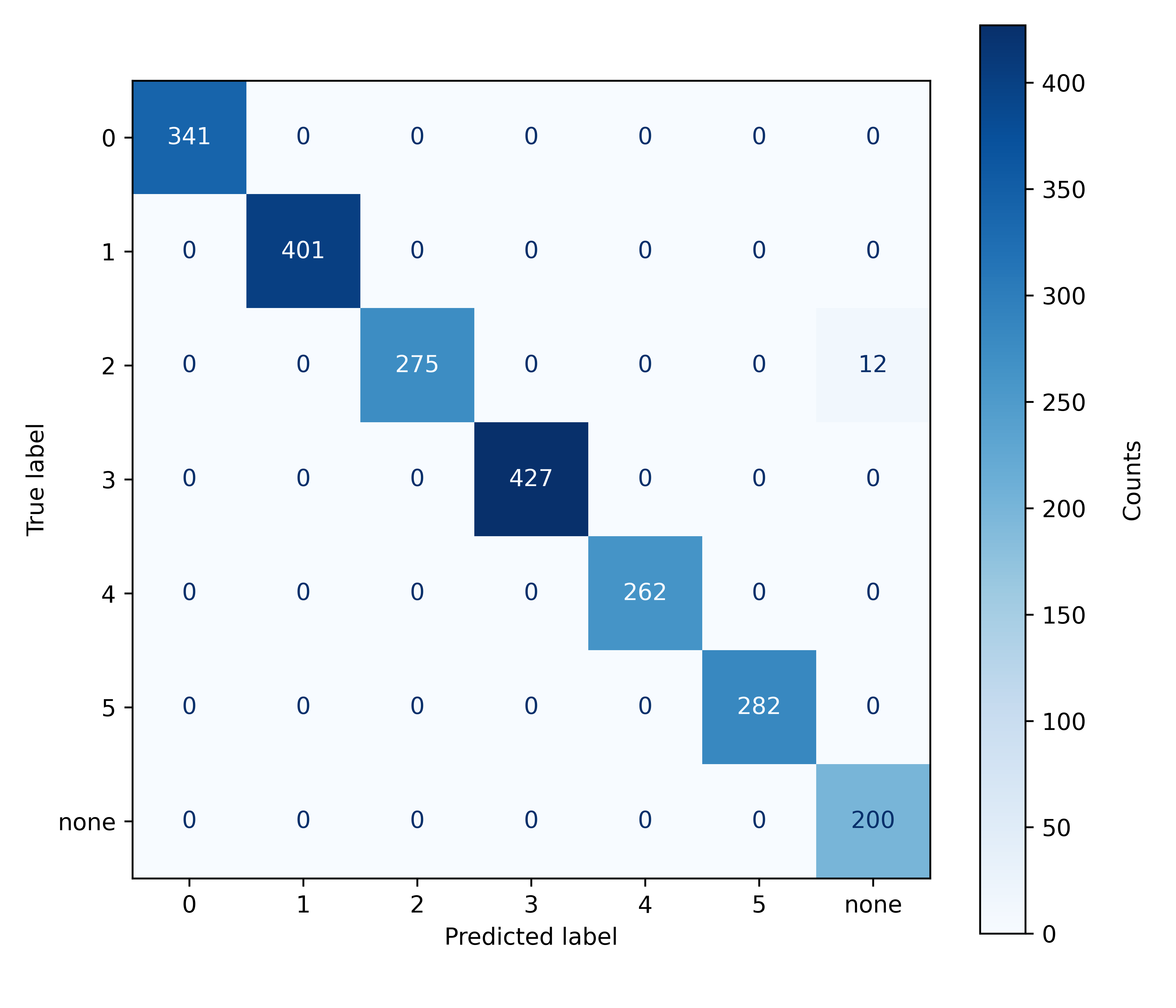}
        \caption{\edit{Results of the classification of the unseen commands with OOD commands.}}
        \label{fig:confusion_mtx}
    \end{subfigure}
    \caption{\edit{Out-of-distribution (OOD) detection using the latent space of text commands (spacecraft proximity operations). }}
    \label{fig:ood_experiments}
\end{figure}

First, Fig.~\ref{fig:txt_umap} illustrates a two-dimensional Uniform Manifold Approximation and Projection (UMAP) \cite{mcinnes2020umap} of the text embeddings associated with the spacecraft proximity operations scenario.
The embeddings form clear clusters aligned with behavior modes across both seen (training) and unseen (test) templates, indicating that the model captures meaningful distinctions in sentence-level semantics. 
This structure naturally supports OOD detection: a user command whose embedding falls far from the known clusters can be flagged as unsupported, preventing unintended or unsafe behaviors. 
Thus, while limited generalization presents challenges, it also provides a stable semantic backbone for real-time supervision.

To further validate the concept of a run-time monitor for OOD command detection, a simple statistical OOD detector is developed based on the Mahalanobis distance computed using a Ledoit--Wolf shrunk covariance estimator \cite{ledoit2004well} in a high-dimensional latent space. 
Given a text embedding $\boldsymbol{e}$, the associated behavior mode is classified according to the following rule:
\begin{subequations}
\begin{align}
    d_b & = (\hat{\boldsymbol{e}} - \bar{\boldsymbol{e}}_b)^\top \Sigma_b^{-1} (\hat{\boldsymbol{e}} - \bar{\boldsymbol{e}}_b), \\
    b^* & = \arg\min_{b \in \mathcal{B}} \; d_b,
\end{align}
\end{subequations}
where $\hat{\boldsymbol{e}} \in \mathbb{R}^{h}$ denotes the mean-pooled text embedding of the input command, and $(\bar{\boldsymbol{e}}_b \in \mathbb{R}^{h}, \Sigma_b \in \mathbb{R}^{h \times h})$ represent the centroid and covariance of the mean-pooled embeddings corresponding to behavior mode $b \in \mathcal{B}$, with $\mathcal{B}$ denoting the set of all behavior modes.

The centroid and covariance associated with each behavior mode are estimated from the input commands in the training dataset using empirical statistics, with the covariance matrix regularized via the Ledoit--Wolf shrinkage estimator to obtain a well-conditioned estimate in the latent space. 
Since the Mahalanobis distance $d_b$ provides a normalized measure of deviation in the latent space, OOD commands can be rejected by thresholding the minimum distance. 
Specifically, a command is classified as OOD if \edit{$d_{b^*} > d_{\min}$}, where $d_{\min}$ is a user-defined threshold.

To evaluate the proposed approach in a spacecraft proximity operations scenario, two dummy behavior modes are additionally defined: (i) \emph{collide with the target spacecraft} and (ii) \emph{grasp and detumble the target spacecraft}.
Using the same procedure described in Appendix A, a total of 200 command templates are generated for these dummy modes. 
These commands are then shuffled together with 2{,}000 unseen input commands used in the statistical analysis, and the corresponding behavior modes are predicted using the proposed classifier.
Figure~\ref{fig:confusion_mtx} summarizes the results in the form of a confusion matrix. 
The classifier perfectly rejects the dummy input commands by exploiting the high-dimensional structure of the embedding space, while achieving near-perfect classification accuracy across the legitimate behavior modes, with minor degradation observed for Behavior~2. 
These results indicate that even a simple statistical classifier can serve as an effective run-time monitor, where user commands whose embeddings lie far from known clusters can be flagged as unsupported, thereby preventing unintended or potentially unsafe behaviors.

Beyond OOD detection, the observed clustering structure also suggests a pathway toward richer forms of run-time assurance. 
Embedding space monitors could be leveraged to (i) identify commands that ambiguously span multiple behavior clusters, (ii) detect semantic drift during execution, or (iii) enforce safety guardrails by constraining the system to remain within predefined safe regions of the embedding space. 
Such mechanisms directly align with recent advances in safety filtering and anomaly detection in AI and robotics. 

\subsection{Towards Broadly Capable Semantic Behavior Models for Spacecraft}

This work represents a first step toward a natural-language interface for spacecraft trajectory generation. 
The preceding sections show that SAGES can generate safe trajectories that remain semantically correct with operator intent, demonstrating the feasibility of \edit{language}-conditioned and constraint-aware autonomy for proximity operations.

However, a central focus of future research must be on \textit{improving generalization} and \textit{providing formal performance guarantees} when such systems are queried outside their training distribution. 
While there is growing evidence that language can serve as a grounded representation to support combinatorial generalization of skills and concepts for robotic tasks \edit{\cite{jang2022bc}}, spacecraft trajectory generation requires a joint understanding of linguistic intent and its corresponding physical realization under complex, often highly nonlinear dynamics. 
Closing this gap remains an open challenge.

A central factor is the construction of the coupled text–trajectory dataset. 
In this study, the training corpus spans only a restricted set of behavior patterns and linguistic templates. 
Although this support can be expanded with greater computational resources and more extensive data generation, the current strategy is inherently bounded. 
\edit{In such regimes, a modular pipeline based on intent classification and ART that is conditioned on the integer behavior mode may provide a practical alternative.}
Empirically, we found that performance degrades rapidly when the system is asked to synthesize trajectories corresponding to semantic behaviors outside the training modes. 
Future work should therefore examine the generalization properties of SAGES-like frameworks in larger data regimes and develop principled methods to ensure reliable behavior beyond the training distribution.
At the same time, while limited generalization presents challenges, the proposed SAGES architecture also provides a stable semantic backbone for real-time supervision, as discussed in the previous subsection. 
This trade-off should be further investigated in the applications so that users can choose the dataset generation strategy based on their specific scenarios. 

Another important challenge relates to compositional reasoning. 
Although the transformer processes text at the token level, supervision is provided only at the trajectory level, offering no explicit incentive to extract or recombine sub-behavior elements. 
As a result, the model tends to learn behavior modes as atomic units rather than developing a structured or compositional semantic space. 
Addressing this limitation will be crucial for enabling the synthesis of novel behavior sequences beyond the training distribution.

Several extensions may help broaden expressiveness while maintaining the safety benefits of a well-structured embedding space:
(i) employing a task-adapted or partially trainable text encoder to capture finer semantic distinctions;
(ii) incorporating a hierarchical policy capable of modeling subtask structure and temporal composition \edit{\cite{takubo2026intent}}; and
(iii) integrating SAGES with a higher-level multimodal reasoning model \cite{foutter2025space-llava} to autonomously generate or refine behavior-level commands.

\section{Conclusion} \label{sec:conclusion}

This paper introduces a novel spacecraft trajectory generation framework capable of interpreting language-driven high-level behavioral commands while satisfying complex nonconvex constraints.
The proposed two-stage architecture \edit{integrates a multimodal transformer conditioned on text and constraint information with a Sequential Convex Programming (SCP) refinement stage}.
The transformer generates a high-quality, \edit{language}-conditioned warm-start trajectory that captures the intended behavior, whereas the SCP finds a nearby feasible trajectory that satisfies nonconvex constraints so that it preserves semantic correctness. 
Numerical experiments demonstrate that the proposed framework produces trajectories with significantly fewer constraint violations than traditional waypoint-based approaches, as validated in two challenging scenarios: hardware experiments on a free-flying robotic platform and a fault-tolerant spacecraft proximity operation with continuous-time constraint satisfaction.
\edit{Overall, the proposed framework represents a foundational step toward a new paradigm of semantic trajectory generation for spacecraft, enabling intuitive and explainable language-based commanding without requiring extensive domain expertise in constraint formulation or waypoint design.}

\section*{Appendix} \label{appendix} 

\subsection{Command-Trajectory Dataset Generation}

\begin{figure}[ht!]
    \centering
    \includegraphics[width=0.9\linewidth]{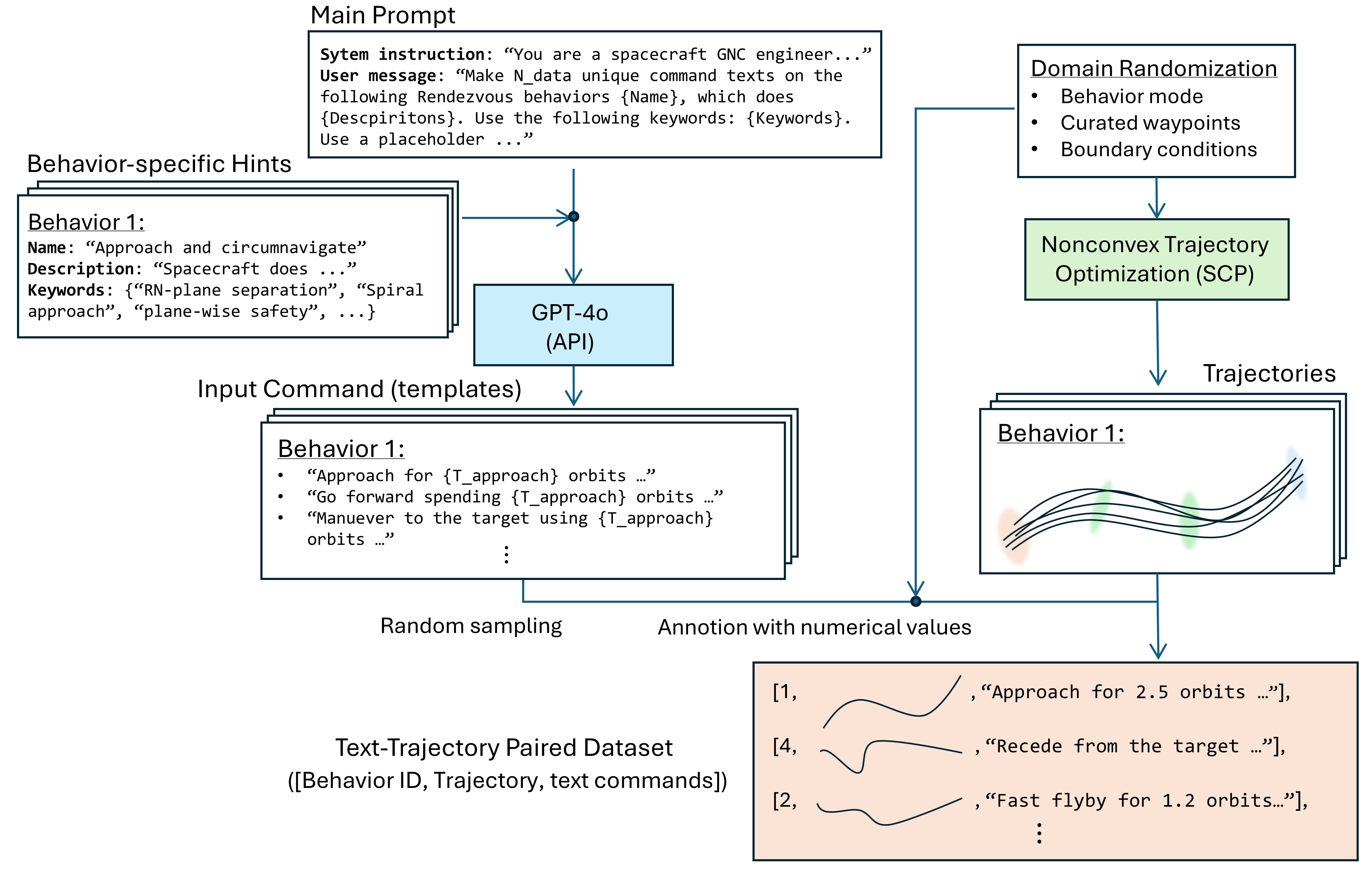}
    \caption{\edit{Flowchart of the dataset generation pipeline}}
    \label{fig:datagen_process}
\end{figure}

This subsection elaborates on the dataset generation process adopted in this paper, summarized in Fig.~\ref{fig:datagen_process}.
To generate a semantically meaningful trajectory, the behavior modes that a spacecraft can take are first enumerated. 
Each behavior mode is represented as a sequence of waypoint regions with associated passage times. 
Using these waypoint specifications, a broad family of problem instances is created through domain randomization of key parameters such as boundary conditions and flight durations. 
For every randomized scenario, a nonconvex trajectory optimization problem is solved via SCP to produce a trajectory sample containing the full history of states, controls, constraint-to-go values, and other relevant parameters.
In this work, to ensure that the dataset contains a sufficient number of feasible trajectories, only problem instances in which the SCP procedure successfully converged from a convex initialization are retained; when converged, both the convex trajectory and its SCP-refined counterpart are stored. 
In parallel, a large pool of input command templates is produced for each behavior mode. 
In this work, these templates are generated automatically using the GPT-4o API \cite{openai_gpt4o_2024}, which synthesizes diverse, behavior-specific command forms based on curated keywords and descriptions with careful prompting. 
If the input command requires numerical values associated with the trajectory's property, numerical placeholders are placed in the command templates, which are later instantiated to produce concrete, scenario-consistent commands.
Finally, trajectories and commands belonging to the same behavior mode are paired by random sampling and stored as elements of the text–trajectory dataset. 

For each behavior mode, 120 distinct text commands (or command templates) are produced, where 100 are allocated to training, and 20 are reserved for testing, where the trajectory and the input command are randomly matched during the dataset generation. 

The following sections present the prompt used to generate text commands for each behavior mode, along with an illustrative example of the associated keywords.

\subsubsection{Free-flyer scenario}

\noindent\textbf{System instruction: }

\begin{lstlisting}
You are an expert GNC technical writer for proximity operations on a microgravity bench. Produce ONE sentence per input describing a goal-directed trajectory with KOZ compliance and left/right/central corridor behavior. Be concise (less than 15 words), precise, and varied in style. Avoid jargon bloat.
\end{lstlisting}

\noindent\textbf{User message template}
\begin{lstlisting}
Behavior description: {beh_text}
Task: Generate ONE sentence that characterizes this specific trajectory mode on a microgravity
testbed, capturing (i) path geometry, (ii) avoidance strategy, (iii) corridor selection,
and (iv) implied execution tempo.
Speed context: {speed_clause}
Style controls: Apply {style_voice}; {style_tone}; {style_structure}.
Guidance: Prefer concise proximity-operations phrasing (e.g., KOZ, standoff, LOS, RCS).
Do not reuse long phrases from the behavior description.
Strict constraints: <15 words; neutral and technically precise; no bullet points; no quotes;
avoid the following terms: {do_not_use}
Vocabulary hints: {geom_hints}; {spd_hints}
\end{lstlisting}

\noindent\textbf{Example for Behavior 0}

\begin{lstlisting}
cmd = "Execute a rapid port-side arc that clears the KOZ while maintaining efficient transit."
\end{lstlisting}

\subsubsection{Spacecraft proximity operation}

\noindent\textbf{System instruction: }

\begin{lstlisting}
You are an expert spacecraft GNC engineer and operator. Generate unique, short imperative templates with placeholders that will be filled later.
\end{lstlisting}

\noindent\textbf{User message:}

\begin{lstlisting}    
RPO Command: {cmd}
Generate {K} unique, short command templates (<={max_words} words) that ask the spacecraft to perform this behavior.
Behavior hints (optional vocabulary): {lexicon_hint}
Behavior structure and placeholder usage: {explanation_hint}
Hard constraints:
 0. Behavior structure: In each template, first state the terminal goal, then describe how it is achieved. Enumerate all phases in the order implied by the explanation; do not reorder phases.
 1. Uniqueness: Every template must be unique. Do not reuse the same sentence pattern.
 2. Length: Each template must be {max_words} words or fewer.
 3. Style: Use imperative commands directed at the spacecraft (e.g., Approach..., Hold..., Flyby...).
 4. Placeholders: Use only the allowed placeholders {ph_str}, written exactly as {name}. Follow the phase order specified in the explanation.
 5. Units: Every placeholder denotes a time in orbits or a distance in meters. When used, append the unit (e.g., {T_appr_orbits} orbits, {d_lambda_meters} m).
 6. Placeholders {T_transfer_orbits} and {T_circ_orbits} denote the epoch at which a phase ends. They must be expressed using structures such as until {placeholder} orbits or ends at {placeholder} orbits. They must never be phrased as durations, such as for {placeholder} orbits.
\end{lstlisting}

\noindent\textbf{Example for Behavior 0: }

\begin{lstlisting}
cmd = "Approach to the relative orbit around the target, and circumnavigate"
ph_str = "T_appr_orbits"
lexicon_hint = "E/I vector separation in relative orbit, spiral approach, plane-wise safety, RN-plane separation, after approach, skirt the keep-out zone"
explanation_hint = "Use {T_appr_orbits} orbits to make a spiral approach; upon arrival, circumnavigate for the remaining time with RN-plane separated safe relative orbit."
\end{lstlisting}

\subsection{Free-flyer scenario Problem Parameters} \label{sec:appendix_ff}

Table \ref{tab_appendix:ff_param} presents the problem-specific parameters used in the free-flyer scenario.
\begin{table}[ht] \small
\centering
\caption{Parameters in the free-flyer scenario.}
\renewcommand*{\arraystretch}{1.2}
\begin{tabular}{cccccccc}
\thickhline
$\edit{\{\theta_0^{(0)}, \theta_0^{(2)}\}} \, [^\circ]$ & $\Delta\theta_0\,[^\circ]$ & 
$\{\rho_{\min},\rho_{\max}\}\,[\mathrm{m}]$ &
$R$ [m] & 
$T/m\,[\mathrm{N/kg}]$ &
$\epsilon_g \,[\mathrm{m}]$ & 
$N$ & $\Delta t\,[\mathrm{s}]$ \\
\hline
\edit{\{122, -122\}}  & $30$ & $\edit{\{0.425, 0.485\}}$ &  $0.385$ & $0.005$ & $0.06$ & $100$ & $0.4$ \\
\thickhline
\end{tabular}
\label{tab_appendix:ff_param}
\end{table}

\subsection{Efficient Computation of Continuous-time Passive Safety under Imperfect burns}\label{app:ct_ps} 

This subsection details the efficient computation of the continuous-time constraint satisfaction defined in Eq.~\eqref{eq:ct_ps_integral}. 
By exploiting the LTV dynamics and the analytical STM, the double integral over time $\tau$ and the impulse fraction $\alpha$ can be computed using a semi-analytical approach: exact integration over $\alpha$ followed by Gauss-Legendre quadrature over $\tau$.

Let $\boldsymbol{x}(t_k; 0, \alpha) = \boldsymbol{x}_k + \alpha \Gamma_k \boldsymbol{u}_k$ denote the state immediately following a fractional impulse $\alpha$. 
The drifted state at time $t_k+\tau$ along the reference state is given by $\bar{\boldsymbol{x}}(t_k; \tau, \alpha) = {\Phi}(t_k+\tau, t_k)\bar{\boldsymbol{x}}(t_k; 0, \alpha)$.
Defining the auxiliary symmetric matrix $\Theta(t_k,\tau)$ and scalar coefficients $a, b, c$ as
\begin{subequations}
\begin{align}
    & \Theta(t_k,\tau) := {\Phi}(t_k+\tau, t_k)^\top S_{k\tau}\,{\Phi}(t_k+\tau, t_k), \\
    & a(\tau) := \bar{\boldsymbol{x}}_k^\top \Theta(t_k,\tau) \bar{\boldsymbol{x}}_k, \quad 
      b(\tau) := \bar{\boldsymbol{x}}_k^\top \Theta(t_k,\tau) \Gamma_k\bar{\boldsymbol{u}}_k , \quad 
      c(\tau) := (\Gamma_k\bar{\boldsymbol{u}}_k)^\top \Theta(t_k,\tau) \Gamma_k\bar{\boldsymbol{u}}_k, \label{eq:abc}
\end{align}
\end{subequations}
the function $\bar{g}(t_k; \tau,\alpha)$ defined in Eq.~\eqref{eq:ct_ps_integral} becomes a concave quadratic in $\alpha$:
\begin{equation} \label{eq:gbar_alpha_quad}
    \bar{g}(t_k; \tau,\alpha) := 1 - \bar{\boldsymbol{x}}(t_k; \tau,\alpha)^\top S_{k\tau}\, \bar{\boldsymbol{x}}(t_k;\tau,\alpha) 
    = 1 - (1 + 2b(\tau))\alpha - c(\tau)\alpha^2.
\end{equation}

For a fixed $\tau$, the constraint violation occurs within the interval $[\alpha_1, \alpha_2] \subseteq [0,1]$ where $\bar{g}(t_k; \tau, \alpha) \geq 0$, which is simply a quadratic inequality. 
The fundamental moments of $\bar{g}(t_k; \tau,\alpha)$ over this active set admit closed-form solutions:
\begin{subequations} \label{eq:alpha_moments}
\begin{align}
I_0(\tau) & =\int_{\alpha_1}^{\alpha_2} \bar{g}(t_k; \tau,\alpha)  d\alpha = (1-a)\Delta_1 - b\Delta_2 - \dfrac{c}{3}\Delta_3, \\
I_1(\tau) & =\int_{\alpha_1}^{\alpha_2} \alpha \bar{g}(t_k; \tau,\alpha)  d\alpha = \dfrac{1-a}{2}\Delta_2 - \dfrac{2b}{3}\Delta_3 - \dfrac{c}{4}\Delta_4, \\
I_2(\tau) & =\int_{\alpha_1}^{\alpha_2}  \alpha^2 \bar{g}(t_k; \tau,\alpha)  d\alpha = \dfrac{1-a}{3}\Delta_3 - \dfrac{b}{2}\Delta_4 - \dfrac{c}{5}\Delta_5,
\end{align}
\end{subequations}
with $\Delta_n := \alpha_2^n - \alpha_1^n$.
Utilizing the chain rule on Eq.~\eqref{eq:convexified_ps}, the gradient terms $G_k^x$ and $G_k^u$ are derived as weighted sums of these moments:
\begin{subequations} \label{eq:G_deriv}
\begin{align} 
G_k^x &= -4\int_{0}^{\tau^s} \Theta(t_k, \tau)\bigl[\bar{\boldsymbol{x}}_k\,I_0(\tau)+\Gamma_k\bar{\boldsymbol{u}}_k\,I_1(\tau)\bigr]\,d\tau, \quad G_k^u = -4\int_{0}^{\tau^s} \Gamma_k^\top \Theta(t_k, \tau)\bigl[\bar{\boldsymbol{x}}_k\,I_1(\tau)+\Gamma_k\bar{\boldsymbol{u}}_k\,I_2(\tau)\bigr]\,d\tau.
\end{align}
\end{subequations}
Similarly, the integral cost $\tilde{g}(\bar{\boldsymbol{x}}_k, \bar{\boldsymbol{u}}_k)$ is obtained by integrating the square of Eq.~\eqref{eq:gbar_alpha_quad}:
\begin{align} \label{eq:g_tilde}
\tilde{g}(\bar{\boldsymbol{x}}_k,\bar{\boldsymbol{u}}_k)
 = \int_{0}^{\tau^s} \left[ a^2\Delta_1
-2 a b \Delta_2
+\left(\dfrac{4}{3} b^2 -\dfrac{2}{3} a c \right) \Delta_3
+ b c \Delta_4
+\dfrac{1}{5} c^2 \Delta_5 \right] d\tau.
\end{align}
Note that the coefficients $a, b, c$ and the roots $\alpha_{1,2}$ are implicit functions of $\tau$.

Since the STM is analytical, the remaining integrals over $\tau$ in Eqs.~\eqref{eq:G_deriv} and \eqref{eq:g_tilde} are efficiently calculated using the Gauss–Legendre quadrature rule. The time interval $\tau \in [0, \tau^s]$ is mapped to the quadrature domain $\xi \in [-1, 1]$ via the transformation $\tau(\xi) = \frac{\tau^s}{2}(\xi + 1)$.
The integral of a generic function $h(\tau)$ is approximated as
\begin{align}
    \int_{0}^{\tau^s} h(\tau) \, d\tau \approx \frac{\tau^s}{2} \sum_{j=1}^{N_q} w_j h\left(\tau(\xi_j)\right), 
\end{align}
where $N_q$ is the number of sample points, and $(\xi_j, w_j)$ are the standard Legendre nodes and weights.
Consequently, the convexified constraints are evaluated by summing the integrands of Eqs.~\eqref{eq:G_deriv} and \eqref{eq:g_tilde} at the $N_q$ evaluation points. 
This approach eliminates the need for dense ODE propagation while maintaining high accuracy for constraint satisfaction, making the constraint formulation more appealing to onboard applications. 
In the spacecraft proximity operation scenario, $N_q=30$ is adopted. 

\subsection{SCP Parameters} \label{sec:appendix_scp}

Table \ref{tab_appendix:scp_param} presents the hyperparameters of the SCP used for the spacecraft proximity operation scenario.
\begin{table}[ht]
\centering
\caption{Hyperparameters of SCVx* \cite{oguri2023successive} used for spacecraft proximity operation scenario.}
\renewcommand*{\arraystretch}{1.2}
\begin{tabular}{cccccc}
    \thickhline
    $\epsilon$ &
    $\{\rho_0, \rho_1, \rho_2\}$ &
    $\{\alpha_1, \alpha_2, \beta, \gamma\}$ &
    $\{ r^{(1)}, r_{\min}, r_{\max} \}$ &
    $\{ w^{(1)}, w_{\max} \}$ &
    \# max.\ iter \\
    \hline
    $10^{-3}$ &
    $\{0.0,\,0.25,\,0.7\}$ &
    $\{2,\,2,\,1.5,\,0.9\}$ &
    $\{0.5,\,10^{-6},\,10\}$ &
    $\{10,\,10^9\}$ &
    100 \\
    \thickhline
\end{tabular}
\label{tab_appendix:scp_param}
\end{table}

\subsection{Transformer Model}
The presented transformer-based trajectory generation is implemented in PyTorch \cite{paszke2019pytorch} and builds on Hugging Face's \texttt{transformers} library \footnote{\url{https://huggingface.co/docs/transformers/index}}.
Table \ref{tab_appendix:hyper} presents the hyperparameter settings used in this work.

\begin{table}[ht]
\centering
\caption{Hyperparameters of the causal transformer.}
\begingroup
\renewcommand*{\arraystretch}{1}
\begin{tabular}{l l}
    \thickhline
     Hyperparameter & Value \\
    \hline
     Number of layers & 6\\
     Number of attention heads & 6 \\
    Embedding dimension, $h$ & 384 \\
     Batch size& 4 \\
    Context length $K$ & 50 \\
    Non-linearity & ReLU\\
    Dropout & 0.1\\
    Learning rate & 3e-5\\
    Grad norm clip & 1.0 \\
    Learning rate decay & None \\
    Gradient accumulation iters & 8 \\
    \thickhline
    \end{tabular}%
    \label{tab_appendix:hyper}
    \endgroup
\end{table}

\section*{Acknowledgements}

This work is supported by Blue Origin (SPO \#299266) as an Associate Member and Co-Founder of Stanford’s Center of AEroSpace Autonomy Research (CAESAR). 
This article solely reflects the opinions and conclusions of its authors and not any Blue Origin entity. 
Yuji Takubo acknowledges financial support from the Ezoe Memorial Recruit Foundation.
AI tools (ChatGPT, Gemini) were used to improve the clarity and grammar of the manuscript.

\bibliography{bibliography}

\end{document}